\pdfoutput=1

\documentclass[11pt]{article}

\usepackage{acl}
\usepackage{times}
\usepackage{latexsym}
\usepackage[T1]{fontenc}
\usepackage[utf8]{inputenc}

\usepackage{microtype}

\usepackage{graphicx}
\usepackage{amsmath}
\usepackage{amssymb}
\usepackage{algorithm}
\usepackage{algpseudocode}
\usepackage{enumitem}
\usepackage{makecell}
\usepackage{array}
\usepackage{url}
\usepackage{multirow}
\usepackage{gensymb}
\usepackage[symbol]{footmisc}

\usepackage{microtype}

\usepackage{multirow}
\usepackage{array}
\newcolumntype{N}{>{\centering\arraybackslash}m{0.75cm}}
\newcolumntype{T}{>{\centering\arraybackslash}m{1.3cm}}
\newcolumntype{S}{>{\centering\arraybackslash}m{1.5cm}}
\newcolumntype{K}{>{\centering\arraybackslash}m{2.2cm}}
\newcolumntype{I}{>{\centering\arraybackslash}m{2.5cm}}
\newcolumntype{M}{>{\centering\arraybackslash}m{3cm}}
\newcolumntype{J}{>{\centering\arraybackslash}m{3.7cm}}
\newcolumntype{L}{>{\centering\arraybackslash}m{4cm}}

\title{Towards Robust and Semantically Organised Latent Representations for Unsupervised Text Style Transfer}

\author{Sharan Narasimhan \hspace{0.2cm}  Suvodip Dey \hspace{0.2cm} Maunendra Sankar Desarkar  \\ \\
  Indian Institute of Technology Hyderabad, India
  \\
 \texttt{\small sharan.n21@gmail.com} \hspace{0.1cm} \texttt{\small suvodip15@gmail.com} \hspace{0.1cm} \texttt{\small maunendra@cse.iith.ac.in} \\ 
}
\date{}

\begin{document}
\maketitle
\begin{abstract}
    Recent studies show that auto-encoder based approaches successfully perform language generation, smooth sentence interpolation, and style transfer over unseen attributes using unlabelled datasets in a zero-shot manner. The latent space geometry of such models is organised well enough to perform on datasets where the style is ``coarse-grained'' i.e. a small fraction of words alone in a sentence are enough to determine the overall style label. 
    A recent study uses a discrete token-based perturbation approach to map ``similar'' sentences (``similar'' defined by low Levenshtein distance/ high word overlap) close by in latent space. This definition of ``similarity'' does not look into the underlying nuances of the constituent words while mapping latent space neighbourhoods and therefore fails to recognise sentences with different style-based semantics while mapping latent neighbourhoods.
    We introduce EPAAEs (Embedding Perturbed Adversarial AutoEncoders) which completes this perturbation model, by adding a finely adjustable noise component on the continuous embeddings space. We empirically show that this (a) produces a better organised latent space that clusters stylistically similar sentences together, (b) performs best on a diverse set of text style transfer tasks than similar denoising-inspired baselines, and (c) is capable of fine-grained control of Style Transfer strength. We also extend the text style transfer tasks to NLI datasets and show that these more complex definitions of style are learned best by EPAAE. To the best of our knowledge, extending style transfer to NLI tasks has not been explored before. \footnote{Our code, data and outputs are available at \url{https://github.com/sharan21/EPAAE}}
\end{abstract}
\section{Introduction}

\begin{figure}[b]
    \centering
    \includegraphics[width=0.46\textwidth]{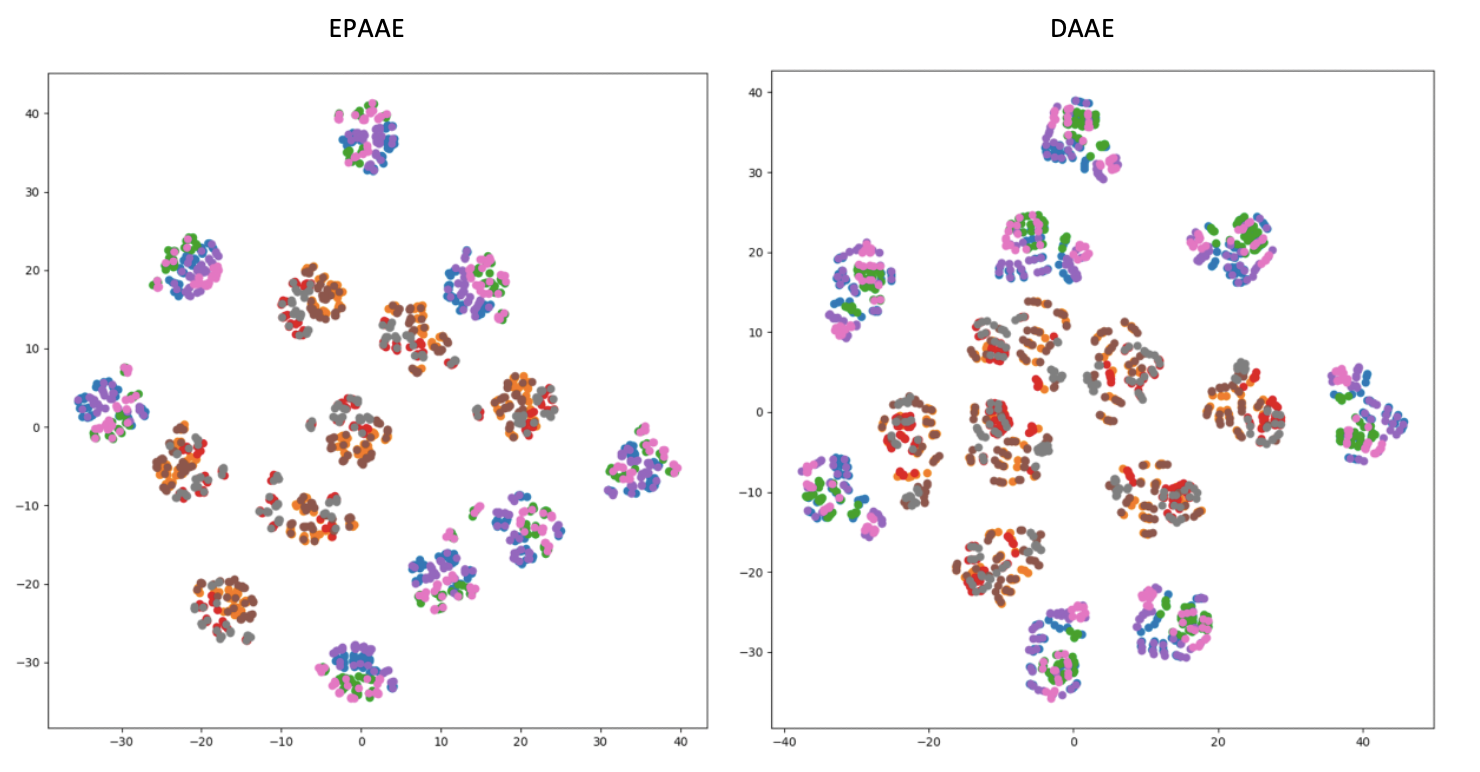}
    \caption{A TSNE plot of encoded latent vectors for all 1950 sentences in the Toy Dataset. EPAAE shows tighter and more organised clustering. Refer to  \ref{fig:fig1big} and \ref{fig:fig2big} for larger plots with legend.}
    \label{fig:fig1}
\end{figure}

The Text Style transfer (TST) task is a form of controlled language generation. The goal is to produce fluent style-altered sentences from a given base sentence, while also preserving its style-independent content. The definition of ``style'' depends on the class labels of the end task. Our work focuses on the unsupervised scenario i.e. performing the training on completely unlabelled corpora. By inducing the latent space organization through input perturbation, TST can be performed using a simple vector arithmetic method (discussed in Section \ref{experiments}).

\textbf{Background.} Several well-known architectures use auxiliary objectives that serve as regularizers to ensure that the latent space geometry is smoothly interpolatable and learns high-level semantic features (refer Appendix \ref{autoencoderwork}). Inspired by successes in denoising approaches in vision \cite{creswell, vincent}, we look at input perturbation based approaches for unsupervised style transfer. Unlike text, in vision, freedom exists to finely control the degree of Gaussian ``blur'' on the continuous input image space. We conjecture that this notion of ``controlling blur'' may be useful in the text domain as well, serving as our motivation for our chosen model for Input Perturbation. \citet{daae}  map ``similar'' (defined by low Levenshtein distance/ high word overlap) sentences together in the latent space by introducing a simple denoising objective over an underlying Adversarial Autoencoder (AAE) \cite{aae}. This noise model includes simple token manipulation (token dropout and substitution) with some probability $p$, to reconstruct the original sentences from the perturbed inputs. We can reason intuitively that discrete word dropout allows sentences with high word overlap (or low Levenshtein distance) to have a higher chance of being perturbed into one another, thereby mapping them close in the latent space during the training time. However, this method allows for stylistically dissimilar sentences, albeit with high word overlap, to be mapped together in the latent space.\\
\textbf{Idea.} We argue that this negatively impacts the quality of final latent geometry and the results of the subsequent style transfer task. 
As an alternative, we explore Embedding Perturbation, where a noise vector is sampled, appropriately scaled and added to the embeddings of each input sentence, such that the resultant embeddings are constrained to live inside an E-dimensional hypersphere. The radius of this hypersphere is controllable using a tune-able hyper-parameter $\zeta$. This hypersphere constraint is partially inspired by concepts in adversarial robustness and ensures that each resultant noised word embedding is not altered to the extent of causing the underlying semantics of the sentence to change. We argue that this allows sentences with stylistically similar constituent embeddings to be mapped together, also encouraging the formation of style-preserving latent neighbourhoods (more on this in Section \ref{latentnn}). The resulting latent representation from ``embedding-perturbed'' autoencoders consequently are better semantically organized and stylistically robust, enabling us to perform TST using an inductive method i.e. simple vector arithmetic on latent vectors.
\\
\textbf{Contributions.} We show that this extended model of input perturbation with both discrete and continuous components, allows for overall better quality text style transfer, particularly in its ability to preserve style-independent ``content'' information.  
To expand the traditional definitions of styles such as Polarity, Formality, which are based on simple ``intra-sentence'' attributes, we introduce the ``Discourse Style Transfer Task'' by salvaging NLI datasets in which the flow of logic between sentences are captured using ``Entailment'', ``Contradiction'', and ``Neutral'' labels. This enables interesting applications such as discourse manipulation in which a pair of sentences agreeing with each other can be made to contradict, and vice versa. We also test our model on fine-grained styles present in the Style-PTB dataset. We empirically show that our model performs the best on a diverse set of datasets with styles ranging from coarse-grained styles (like sentiment) to fine-grained styles (like tenses) and complex inter-sentence styles (like discourse or flow of logic).

 \section{Related work}

\textbf{Seminal work in TST. } Autoencoder based approaches for TST on labelled non-parallel datasets are quite well explored \cite{shen, fu}. Some techniques involve implicit Style-Content disentanglement of the latent space using Back Translation \cite{backtranslation} and adversarial learning \cite{john}. \citet{simple} achieve disentanglement using simple keyword replacement. Most studies look at simple non-parallel classification datasets, defining their style to be the class label. Studies also look at Syntax-Semantic disentanglement of the latent space \cite{syntax1, syntax2}. A $\lambda_{1}$ penalty is imposed on the log variance of the perturbations to prevent it from vanishing. The latent vacancy problem \cite{cpvaee} of the $\beta-$VAE is mitigated by introducing auxiliary losses and provided for one of the earliest methods for unsupervised TST.
 Similar to our work, \citet{laae} introduces the Latent noised AAE (LAAE), Gaussian perturbation is instead added to latent encodings to promote organization. Unsupervised work includes using a language model as a discriminator for a richer feedback mechanism \cite{lang}, allowing it to increase performance in word substitution decipherment, sentiment modification, and related language translation. More seminal work related to autoencoders in the context of Style Transfer is mentioned in Appendix \ref{autoencoderwork}. \\
 \textbf{Contemporary work in TST. }
 More recent work, treat the style transfer problem as paraphrase generation and fine-tune pre-trained language models \cite{paraphrase}. \citet{padded} trains masked language models or MLMs on the source and target domains to identify input tokens to be removed and replace them with tokens from the target MLM in an unsupervised manner. \citet{aaai} uses gradient-based update rules in the continuous latent space $z$ from style and content predictor networks, enabling the transfer of fine-grained styles without using an adversarial approach. \citet{lewis} performs TST on sentiment and politeness datasets using token editing methods (similar to \citet{simple}) using Levenshtein editing operations. \citet{enhancingcp}
also focuses on enhancing content preservation by introducing a method to remove style at the token level using reverse attention and fuse this content representation with style using a conditional layer normalization technique. \citet{textsetter} adapts the T5 model \cite{t5} for few shot text style transfer by extracting a style representation and performing conditioned decoding, using only a handful of examples at inference time.

\section{Method}

The underlying language model is a generative auto-encoder which models an input distribution $p(\mathbf{x})$ assumed to be from an underlying latent distribution $p(\mathbf{z})$. A deterministic encoder $E$ representing the distribution $q( \mathbf{z} | \mathbf{x})$, in the form of an RNN, whose output is reparameterized by another distribution $z_i \sim  \mathcal{N}(\mu_i(\mathbf{x}),\sigma_i(\mathbf{x}))$ to give the aggregated posterior distribution $q(\mathbf{z})$. Various auxiliary loss functions are used to enforce the learned prior $q(\mathbf{z})$ to match $p(\mathbf{z})$.
The Generator $G$ representing $p( \mathbf{x} | \mathbf{z})$, also in the form of an RNN, decodes back the sample from the learnt prior $q(\mathbf{z})$ into its corresponding input from $p(\mathbf{x})$. Gradient descent is applied on the reconstruction loss of the autoencoder given by:
\begin{equation}
\mathcal{L_{\text{rec}}}(\theta_E,\theta_G) = E_{p_{\text{data}}(x)}[-\log p_G(x|E(x))] 
 \label{eq:rec}
\end{equation}

We use the AAE \cite{aae} as our choice for the underlying generative autoencoder over which the input perturbation techniques were applied. AAE uses a discriminator $D$ to enforce $q(\mathbf{z})$ to match $p(\mathbf{z})$, a standard Gaussian, by learning to distinguish between samples from the two different distributions. This adversarial loss serves as a regularizer for the latent space, giving it the ability the organize itself better for smooth sentence interpolation. 
\begin{align}
    \mathcal L_{\text{adv}}(\theta_E,\theta_D) =~ &  \mathbb E_{p(z)}[-\log D(z)] ~+ \nonumber \\
    & \mathbb E_{p_{\text{data}}(x)}[-\log(1-D(E(x)))]
\end{align}

The final min-max objective is a sum of the reconstruction loss (given below) and $\lambda$ weighted adversarial loss:
\begin{align}
    \min_{E,G}\max_{D}  ~ \mathcal L_{\text{rec}}(\theta_E,\theta_G) - \lambda \mathcal L_{\text{adv}}(\theta_E,\theta_D)\label{eq:aae}
\end{align}

We found empirically that AAEs performed well and were stable during training. On the other hand, $\beta$-VAEs required careful tuning of the $\beta$ hyperparameter to prevent posterior collapse and did not perform as well.

\subsection{Finely-controlled continuous noise on embedding space}

\begin{figure}[htb]
    \centering
    \includegraphics[width=0.45\textwidth]{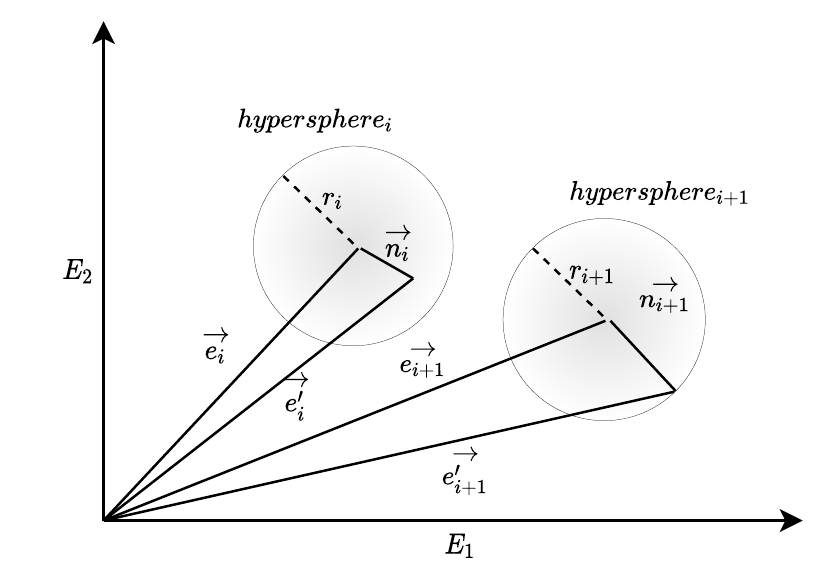}
    \caption{An abstract representation of the continuous embedding noise approach for two arbitrary embeddings $e_i$ and $e_{i+1}$ on an embedding space $E\in R^2$}
    \label{fig:fig2}
\end{figure}

To further organize the latent geometry of the underlying AAE to encode style-based semantic similarity among sentences, we propose a perturbation-based approach on the continuous embedding space. Word embeddings of dimensionality $E$, of each input token $x_{i}$ are denoted as $e_{i}$. Consider an input sentence of length $l$ containing tokens $x_{0}, \cdots, x_{l-1}$ having embeddings $e_{0}, \cdots, e_{l-1}$ respectively. Our objective is to blur every embedding vector $e_{i}$ by adding some appropriately scaled noise vector $n_{i}$ to produce a resultant noised embedding vector $e'_{i}$, such that $e'_{i}$ does not lie too far away from $e_{i}$ to not change the underlying semantics of the word completely. We do this by ensuring that each new $e'_{i}$ lives inside an $E$-dimensional hypersphere. The center of the hypersphere is the original embedding $e_i$ and its radius is defined as $|e_{i}|$ * ${s_{i}}$, where $ {s_{i}}\in \mathbb{R}^1$ is a random variable sampled from a distribution $P(s)$. $P(s)$  is probability distribution of the form $ \mathcal {Y}(\mu=0,\sigma)$ where ${\mathcal{Y}}$ is some arbitrarily chosen distribution and $\sigma$ as function of hyperparameter $\zeta \in R$. This distribution $P(s)$ parameterized by $Y, \zeta)$ models the probability density cloud inside the embedding hypersphere, controlling its radius and interior densities. Figure \ref{fig:fig1} neatly summarizes the aforementioned embedding perturbation mechanism for a simple example with only two individual word embeddings. In practice, we use a vectorized representation of the above mechanism to blur the embeddings of a mini-batch of sentences of size $L$ in constant O(1) time. The embedding perturbation method is summarized below in a vectorized form:

\begin{equation}
\begin{gathered}
    z  \sim \mathcal{N}(\mu,\,\sigma^{2})\, \\ 
    \hat{z} = z/|z| \\
    s \sim Y(\mu=0, \sigma^{2} = {(\zeta/3)}^2) \\
    n = (s\odot|e|)^T\hat{z} \\
    e' = e+n \\
    r = \zeta * |e|
\end{gathered}
\end{equation}

where $z$, $\hat{z}$, $s$, $n$ and $e$ are vectorized representations of $z_{i}$, $\hat{z_{i}}$, $s_{i}$, $n_{i}$ and $e_{i}$ respectively for a mini-batch of sentences of size $l$. $\odot$ and * denote element wise and scalar multiplication respectively. $|x|$ denotes the magnitude of a vector/batch of vectors $x$. $r$ is the vectorized mini-batch representation of $r_{i}$, the expected radius of $hypersphere_{i}$. We choose the Gaussian distribution as our choice of the probability distribution $\mathcal{Y}$, as on testing it produced the best results\footnote{We also tried the uniform distribution with varying values for $\zeta$}. To couple the radius $r_{i}$ of $hypersphere_{i}$ to our hyperparameter $\zeta$ to enable fine-grained control using $\zeta$, we constrain the probability density of $Y$ to live inside the hypersphere within three standard deviations. To achieve this, we equate $3\sigma$ to $\zeta$, and consequently set the variance $\sigma^2$ of $\mathcal{Y}$ to be ${(\zeta/3)}^2$.

\subsection{Discrete word dropout probability}
The ``Denoising Autoencoder'' or DAAE \cite{daae} considered discrete token-level perturbations such as token masking, substitution and deletion. We consider token deletion as discrete noise. Drop probability $p=0.3$ is found to be the best in both their experiments and ours. 

Token deletion is the only type of input perturbation that can alter the number of tokens in a sentence. Any continuous model for input noise cannot mathematically generalize the effects of this kind of discrete word dropout. Furthermore, during our experiments, we find that for some datasets, both discrete and continuous noise components are required to produce the overall best model. In such a case, we first perform token deletion and then subject the leftover token embeddings to perturbation. We refer to this generalized noise model as ``Embeddings-Perturbed Adversarial Autoencoder'' or ``EPAAE'' parameterized by $\zeta$ and $p$.

\section{Semantic Similarity in Latent Space Neighbourhoods}
\label{latentnn}
We contrast and compare the resultant latent space neighbourhoods of DAAEs and EPAAEs. 
 
\subsection{Preliminary reasoning}
Here, we first investigate the question - \textit{Does token deletion during training group truly put semantically similar sentences together in latent space $Z$?}.
Intuitively we can reason that the answer may be in the affirmative if the drop probability is small. For example, the sentences ``The food was good'' and ``The food was superb'' might get perturbed into a common version, i.e. ``The food was'' and therefore be mapped nearby in $Z$ during training. As analysed and concluded in \cite{daae}, latent neighbourhoods in the latent space of DAAEs successfully cluster sentences with high word overlap (low Levenshtein/Edit distance) together. However, the Levenshtein distance metric is not an accurate measure of the true semantic similarity between sentences. A pair of sentences with high word overlap might convey different ideas w.r.t the underlying style-based semantics of the dataset. For example, in the context of the Yelp dataset, the sentences ``The food was good'' and ``The food was bad'' are stylistically opposing (style being polarity in this case) and yet still get mapped close by in Z for DAAEs. 

\subsection{Testing the hypothesis with a Toy Dataset}
 We conduct specifically curated experiments on a synthetic dataset to verify our hypothesis that EPAAEs map stylistically similar sentences together.
 
\textbf{Details of Toy dataset:}  Inspired by Yelp, each sentence in this dataset either represents a positive or negative sentiment. It also contains different sentiment independent components, such as the identity of the person and the subject of the review. Each sentence is of the format: ``The \textit{\textless identity\_token\textgreater} said the \textit{\textless subject\_token\textgreater} is \textit{\textless decision\_token\textgreater}" where
 \textit{identity}, \textit{subject} and \textit{decision} token classes are the only variable parameters in each sentence. The entire set of all permutations of these token classes forms the dataset. For example, the decision class is further subdivided into two subclasses, i.e. positive/ negative sentiment, each subclass containing 7 and 8 choosable tokens, respectively. The resultant dataset consisted of 1950 sentences. The details of the subclasses and the representative tokens inside each subclass are shown in Table \ref{tab:synthd}. 
 We produce output labels for each sentence by using a 3-bit representation, one bit for each of three token classes, where the values of each bit represent the token label subclass within that class. For example, a label of 5 is encoded as 101 corresponding to a sentence with \textit{Female}, \textit{Food} and \textit{Negative} labels.
 There are $2^3$ = 8 labels in total labelled 0-7.
 
\begin{table}
\begin{small}
\centering
\begin{tabular}{SSJ} 
\hline
\textbf{Token Class}  & \textbf{Sub Classes}   & \textbf{Token Options} \\ \hline
\multirow{2}{*}{\textbf{(0) Identity}} 
& 
(0) Male & {boy, man, husband, boyfriend, waiter} \\  
&
(1) Female & {girl, woman, wife, girlfriend, waitress} \\ \hline
\multirow{2}{*}{\textbf{(1) Subject}} 
&
(0) Food & {food, meal, dinner, breakfast, lunch, pasta, chicken} \\ 
&
(1) Others & {service, staff, experience, atmosphere, location, place} \\ 
\hline
\multirow{2}{*}{\textbf{(2) Decision}} 
& 
(0) Positive & {good, great,  excellent, decent, amazing, wonderful, reasonable} \\ 
&
(1) Negative & {bad, worst, horrible, spicy, bland, expensive, disgusting, mediocre}  \\ 
\hline
\end{tabular}
\caption{Structure of token classes and division of sub-classes with token options used to create Toy Dataset}
\label{tab:synthd}
\end{small}
\vspace{-.05in}
\end{table}

\begin{table}
\begin{small}
\centering
\label{tab:toy-nn}
\begin{tabular}{KNKN} 
\hline
\textbf{DAAE ($p=0.3$)} & \textbf{L2 Norm} & \textbf{EPAAE ($\zeta=3.0$)} & \textbf{L2 Norm} \\ 
\hline
the man said the pasta is spicy & 0 & the man said the pasta is spicy & 0 \\ \hline
\textcolor{red}{the man said the pasta is good} & 2.79 & the man said the staff is spicy & 2.62 \\ \hline
the girl said the pasta is spicy & 2.88 & the man said the pasta is bad & 2.69 \\ \hline
\textcolor{red}{the man said the pasta is wonderful} & 2.89 & the man said the pasta is bland & 2.77 \\ \hline
the man said the service is spicy & 3 & the girl said the pasta is spicy & 2.84 \\ \hline
the man said the pasta is bad & 3.02 & the husband said the pasta is spicy & 2.89 \\ \hline
\end{tabular}
\caption{\small Nearest Neighbour experiment performed on EPAAE and DAAE. Sentences in red indicate dissimilarity in semantics with respect to cluster center}
\label{tab:toy-nn}
\end{small}
\end{table}
 
\textbf{Qual. Analysis of Latent space.}  We consider two models, DAAE with token deletion probability $p=0.3$ and EPAAE with $\zeta=2.5$. Both models are initially pre-trained over the unlabelled Yelp dataset. The synthetic dataset is used during inference time only. We encode all 1950 sentences into their respective latent space vectors and use TSNE plots to visualise the latent space of each model (Figure \ref{fig:fig1}). We pick a random \textit{query} sentence, e.g. ``The man said the pasta is spicy'', and encode it into $Z$. We then retrieve the top five nearest encoding to the query and observe their decoded outputs to check for the preservation of style-based semantics (Table \ref{tab:toy-nn}) around the query. We find that the EPAAE maps the latent neighbourhood such that stylistically similar (positive/negative sentiment in this context) are grouped. This is not the case with the DAAE, evident from Table \ref{tab:toy-nn} in which neighbours 1 and 3 have a differing sentiment from the query. This offers an explanation as to why the DAAE is not able to produce tightly confined clusters. 

\textbf{Quant. Analysis of Latent space.} To further validate our hypothesis that EPAAEs better preserve style-based semantics in latent neighbours, we generate useful quantitative metrics over a k-nearest neighbours experiment, done on the test split across all datasets. We document these metrics in Table \ref{latentquant}. Column 3 conveys the mean distance to the closest neighbour with the opposite label. Column 4 conveys the mean number of hops to reach to the closest neighbour with opposite labels.
In all but a few datasets, the DAAE reports a smaller ``mean hops for label flip'', supporting our hypothesis that stylistically dissimilar sentences are mapped closer together in the latent neighbour than our proposed model. We also see this trend mostly holds true for the "Mean L2 Norm for Label Flip" metric in Column 4 as well, providing further evidence to our hypothesis. 
\begin{table}
\begin{small}
\centering
\begin{tabular}{STSS} 
\hline
\textbf{Dataset} & \textbf{Model} & \textbf{Mean L2 Norm for Label Flip} & \textbf{Mean Hops for Label Flip}  \\ \hline
\multirow{2}{*}{Toy Dataset} & DAAE & \textbf{5.725} & 14.979 \\ 
& EPAAE & 4.324 & \textbf{19.587} \\ \hline
\multirow{2}{*}{Yelp}  & DAAE  & \textbf{9.386} & 5.850 \\  
& EPAAE & 8.211 & \textbf{6.513} \\ \hline
\multirow{2}{*}{SNLI} & DAAE & 9.794 & 3.185 \\ 
& EPAAE & \textbf{10.373}  & \textbf{3.528}                                                                                            \\ 
\hline
\multirow{2}{*}{DNLI}        & DAAE           & 9.320                                                                                                        & \textbf{6.027}                                                                                            \\ 

                             & EPAAE          & \textbf{11.681}                                                                                              & 4.882                                                                                                     \\ 
\hline
\multirow{2}{*}{SciTail}     & DAAE           & 10.224                                                                                                       & 5.750                                                                                                     \\ 

                             & EPAAE          & \textbf{10.676}                                                                                              & \textbf{6.253}                                                                                            \\ 
\hline
\multirow{2}{*}{PP Removal}  & DAAE           & 8.195                                                                                                        & 6.722                                                                                                     \\ 

                             & EPAAE          & \textbf{8.774}                                                                                               & \textbf{7.409}                                                                                            \\ 
\hline
\multirow{2}{*}{Tenses}      & DAAE           & 7.686                                                                                                        & \textbf{8.802}                                                                                            \\ 

                             & EPAAE          & \textbf{10.924}                                                                                              & 5.426                                                                                                     \\ 
\hline
\multirow{2}{*}{Voices}      & DAAE           & 6.805                                                                                                        & 8.367                                                                                                     \\ 

                             & EPAAE          & \textbf{9.657}                                                                                               & \textbf{8.726}                                                                                            \\
\hline
\end{tabular}
\caption{Quantitative metrics to capture model's ability to preserve style-based semantics inside Latent Neighbourhoods. Bold indicates model with higher value for that corresponding metric in that specific dataset, which is suggestive of ability to better preserve style-based semantics.}
\label{latentquant}
\end{small}
\vspace{-.2in}
\end{table}

\section{Setup, Datasets and Metrics}\label{sec:tasks_datasets}

\subsection{Experimental Setup}
\textbf{Baselines.} This work focuses on simple denoising approaches for Unsupervised TST and subsequently constrains our choice of baselines to follow these criteria. We consider three other autoencoder based models for our experimentation, i.e. Denoising AAE (DAAE), Latent-Noised AAE (LAAE) and the $\beta$-VAE. 

\textbf{Hyperparameters and Setup.} Details on hyperparameter selection can be found in \ref{hyperparameterselection}.
Other common hyperparameters (detailed in \ref{modelarchitecture}) related to encoder/decoder architectures remained identical across all models. Training is completely unsupervised, and labels are only used during inference time. Details on computation time, number of parameters and infrastructure used can be found in Appendix. \ref{infra}.

\subsection{Datasets}
\label{sub:datasets}
In this section, we briefly describe the different kinds of datasets used for experimentation. Further details is provided in in Appendix \ref{app:datasets}. 

\textbf{Complexity of Styles in Datasets: } 
Current studies mainly focus on high-level styles to validate the approaches. To validate our hypothesis that the EPAAE can perform fine-grained style transfer due to semantics learnt from embeddings, we consider three tasks: sentiment, discourse and fine-grained text style transfer. As prepossessing, we remove non-essential special characters and lowercase all sentences. Except for the Yelp dataset, no pruning is done based on sentence length.

\textbf{Sentiment Style Datasets: } We use the preprocessed version from \cite{shen} of the Yelp dataset. The sentiment labels (positive, negative) were considered as style.

\textbf{Discourse Style Datasets: } To check for the model's ability to alter the discourse or \textit{flow of logic} between two sentences we make use of NLI datasets such as SNLI, DNLI, and Scitail. Each instance in the SNLI dataset \cite{snli} consists of two sentences that either contradict, entail (agree) or are neutral towards each other. Similarly, the DNLI dataset \cite{dnli} consists of contradiction, entailment and neutrality labelled instances. Scitail \cite{scitail} is an entailment dataset created from multiple-choice science exams and the web, in a two-sentence format similar to SNLI and DNLI. The first sentence is formed from a question and answer pair from science exams and the second sentence is either a supporting (entailment) or non-supporting (neutrality) premise.

\textbf{Fine-grained Style Datasets: }
The Style-PTB dataset \cite{styleptb} consists of 21 styles/labels with themes ranging from syntax, lexical, semantic and thematic transfers as well as compositional transfers which consist of transferring more than one of the aforementioned fine-grained styles. To check whether the EPAAE can capture fine-grained styles better by leveraging its better organised latent space, we make use of three styles i.e. Tenses, Voices (Active or Passive) and Syntactic PP tag removal (PPR). In the Tenses dataset, each sentence is labelled with ``Present'', ``Past'', and ``Future''. The Voices dataset contains ``Active'' and ``Passive'' voices labels and the PPR dataset contains ``PP removed'' and ``PP not removed'' labels.

\subsection{Automatic Evaluation metrics}
Evaluation for text style transfer includes checking for a) Style Transfer accuracy, b) Content preservation metrics and c) Fluency of output sentences. Recent studies show that automatic Evaluation metrics are still an open problem and can be gamed \cite{cpvaee}. 

\textbf{Style Transfer Measure:}
A pre-trained classifier is used to check the presence of the target label in the output sentence. \cite{eval} introduces the notion of checking the style transfer intensity apart from just the presence of the target label. While we find this notion intriguing, we wish to first accomplish the style transfer task convincingly for the current set of tasks before assuming a more complex metric. 

\textbf{Content Preservation:}
In recent work, we observe that models typically struggle more in content preservation and the ability to preserve the contextual meaning of the base sentence. The BLEU score alone does not suffice to correlate strongly with actual qualitative results. To truly validate our hypothesis that EPAAE's are better able to preserve content better, we augment the bucket of standard content preservation metrics \footnote{To generate all our content preservation metrics, we use the nlg-eval package from: \url{https://github.com/Maluuba/nlg-eval}}. 
Apart from the standard of using BLEU between sentences of source and target styles, we borrow evaluation techniques from fields similar to Text Style Transfer such as Machine Translation and Text Summarization, such as METEOR \cite{meteor}, ROUGE-L \cite{rogue}, CIDEr \cite{cider} which have been shown to correlate more strongly with human judgement. Following the study of \cite{nlgeval}, in which they show BLEU does not necessarily correlate with human evaluations in dialogue response generation, we also adopt Embedding Average, Vector Extrema \cite{forg} and Greedy matching score \cite{rus}. 

\textbf{Fluency of generations:} Past work measures the perplexity using a pre-trained language model to gauge the fluency or grammatical correctness of the style transferred outputs. \cite{eval} argues such perplexity calculations for style transfer tasks may not necessarily correlate with a human judgement of fluency. Adversarial classifiers in the form of logistic regression networks are trained with the goal of distinguishing between human-produced and machine produced sentences. These classifiers are then used to score the naturalness of the output sentences. We follow this metric during our evaluation of fluency or naturalness \footnote{We use code and pretrained models from \url{https://github.com/passeul/style-transfer-model-evaluation} to measure naturalness.}

\section{Experiments}
\label{experiments}
\begin{table*}[ht!]
\begin{tiny}
\centering
\begin{tabular}{lccccccccc} 
\hline
\textbf{Model}              & \multicolumn{1}{c}{\textbf{Naturalness}} & \multicolumn{1}{c}{\textbf{TST Acc.}} & \multicolumn{1}{c}{\textbf{BLEU-2}} & \multicolumn{1}{c}{\textbf{METEOR}} & \multicolumn{1}{c}{\textbf{ROUGE-L}} & \multicolumn{1}{c}{\textbf{CIDER}} & \multicolumn{1}{c}{\begin{tabular}[c]{@{}l@{}}\textbf{Embedding} \\\textbf{Average}\end{tabular}} & \multicolumn{1}{c}{\begin{tabular}[c]{@{}l@{}}\textbf{Vector} \\\textbf{Extrema}\end{tabular}} & \multicolumn{1}{c}{\begin{tabular}[c]{@{}l@{}}\textbf{Greedy} \\\textbf{Matching} \\\textbf{Score}\end{tabular}}  \\ 
\hline

$\beta$-VAE ($ \beta=0.15$) & 0.792                                     & \textbf{0.819 }                     & 0.064                               & 0.079                                & 0.212                               & 0.271                               & 0.773                                                                                              & 0.445                                                                                           & 0.600                                                                                                              \\ 
LAAE ($\lambda_1=0.05$)     & 0.763                                     & 0.782                               & 0.056                               & 0.072                                & 0.182                               & 0.232                               & 0.756                                                                                              & 0.436                                                                                           & 0.583                                                                                                              \\ 
DAAE ($n=0.3$)              & 0.711                                     & 0.812                               & 0.167                               & 0.134                                & 0.339                               & 0.775                               & 0.810                                                                                              & 0.531                                                                                           & 0.674                                                                                                              \\ 
EPAAE ($\zeta=2.0$)             & \textbf{0.822 }                           & 0.749                               & 0.142                               & 0.125                                & 0.320                               & 0.609                               & 0.800                                                                                              & 0.491                                                                                           & 0.651                                                                                                              \\ 
EPAAE ($\zeta=2.0, p=0.3$)       & 0.708                                     & 0.808                               & 0.193                               & 0.145                                & 0.368                               & 0.838                               & 0.824                                                                                              & 0.542                                                                                           & 0.688                                                                                                              \\ 
EPAAE ($\zeta=2.0, p=0.1$)       & 0.718                                     & 0.771                               & \textbf{0.218 }                     & \textbf{0.170 }                      & \textbf{0.396 }                     & \textbf{1.017 }                     & \textbf{0.827 }                                                                                    & \textbf{0.571 }                                                                                 & \textbf{0.707 }                                                                                                    \\
\hline

\end{tabular}
\caption{Quantitative results of TST experiments on Yelp Dataset}
\label{tab:sentiment}
\vspace{-.1in}
\end{tiny}
\end{table*}

\begin{table*}
\resizebox{1.0\textwidth}{!}{
\centering
\begin{tabular}{ccccccccccc} 
\hline
\textbf{Dataset} & \textbf{Model} & \multicolumn{1}{c}{\textbf{Naturalness}} & \multicolumn{1}{c}{\textbf{TST Acc.}} & \multicolumn{1}{c}{\textbf{BLEU-2}} & \multicolumn{1}{c}{\textbf{METEOR}} & \multicolumn{1}{c}{\textbf{ROUGE-L}} & \multicolumn{1}{c}{\textbf{CIDER}} & \multicolumn{1}{c}{\begin{tabular}[c]{@{}l@{}}\textbf{Embedding }\\\textbf{Average}\end{tabular}} & \multicolumn{1}{c}{\begin{tabular}[c]{@{}l@{}}\textbf{Vector}\\\textbf{~Extrema}\end{tabular}} & \multicolumn{1}{c}{\begin{tabular}[c]{@{}l@{}}\textbf{Greedy }\\\textbf{Matching }\\\textbf{Score}\end{tabular}}  \\ 
\hline
\textbf{SNLI}    & $\beta$-VAE ($ \beta=0.15$)         & 0.983                                     & \textbf{0.523}                      & 0.424                               & 0.257                                & 0.541                               & 2.288                               & 0.948                                                                                              & 0.615                                                                                           & 0.814                                                                                                              \\ 
                 & LAAE ($\lambda_1=0.05$)        & 0.946                                     & 0.522                               & 0.036                               & 0.051                                & 0.158                               & 0.026                               & 0.824                                                                                              & 0.318                                                                                           & 0.593                                                                                                              \\ 
                 & DAAE ($p=0.3$)          & 0.980                                     & 0.519                               & 0.415                               & 0.254                                & 0.534                               & 2.062                               & 0.950                                                                                              & 0.650                                                                                           & 0.838                                                                                                              \\ 
                 & EPAAE ($\zeta=2.5$)          & 0.979                                     & 0.480                               & \textbf{0.534}                      & \textbf{0.316}                       & \textbf{0.655}                      & \textbf{3.615}                      & \textbf{0.958}                                                                                     & \textbf{0.688}                                                                                  & \textbf{0.857}                                                                                                     \\ 
                & EPAAE ($\zeta=2.5,p=0.3$)    & \textbf{0.983}                            & 0.513                               & 0.388                               & 0.240                                & 0.504                               & 1.783                               & 0.948                                                                                              & 0.626                                                                                           & 0.826                                                                                                                \\ 

 & EPAAE ($\zeta=2.5,p=0.1$)    & 0.978                                     & 0.511                               & 0.461                               & 0.281                                & 0.582                               & 2.492                               & 0.953                                                                                              & 0.674                                                                                           & 0.854  
                                                                                                                           \\ 
\hline 

\textbf{DNLI}    & $\beta$-VAE ($ \beta=0.15$)         & 0.927                                     & 0.578                               & 0.243                               & 0.144                                & 0.356                               & 0.654                               & 0.905                                                                                              & 0.503                                                                                           & 0.722                                                                                                              \\ 
                 & LAAE ($\lambda_1=0.05$)        & 0.935                                     & \textbf{0.604}                      & 0.205                               & 0.128                                & 0.319                               & 0.384                               & 0.897                                                                                              & 0.475                                                                                           & 0.704                                                                                                              \\ 
                 & DAAE ($p=0.3$)          & 0.933                                     & 0.597                               & 0.416                               & 0.237                                & 0.522                               & 2.269                               & 0.934                                                                                              & 0.616                                                                                           & 0.803                                                                                                              \\ 
                 & EPAAE ($\zeta=2.5$)          & \textbf{0.939}                                     & 0.513                               & 0.440                               & 0.253                                & 0.557                               & 2.573                               & 0.943                                                                                              & 0.635                                                                                           & \textbf{0.818}                                                                                                     \\ 
                 & EPAAE ($\zeta=2.5,p=0.3$)    & 0.927                                     & 0.574                               & 0.397                               & 0.225                                & 0.506                               & 2.064                               & 0.933                                                                                              & 0.606                                                                                           & 0.798                                                                                                              \\ 
                 & EPAAE ($\zeta=2.5,p=0.1$)    & 0.934                            & 0.579                               & \textbf{0.457}                      & \textbf{0.260}                       & \textbf{0.559}                      & \textbf{2.912}                      & \textbf{0.941}                                                                                     & \textbf{0.636}                                                                                  & 0.636                                                                                                              \\ 
\hline
\textbf{Scitail} & $\beta$-VAE ($ \beta=0.15$)         & 0.770                                     & \textbf{0.570}                      & 0.095                               & 0.065                                & 0.173                               & 0.142                               & 0.796                                                                                              & 0.387                                                                                           & 0.643                                                                                                              \\ 
                 & LAAE ($\lambda_1=0.05$)        & 0.832                                     & 0.497                               & 0.196                               & 0.124                                & 0.287                               & 0.752                               & 0.874                                                                                              & 0.489                                                                                           & 0.712                                                                                                              \\ 
                 & DAAE ($p=0.3$)          & 0.792                                     & 0.454                               & 0.325                               & 0.199                                & 0.413                               & 1.560                               & 0.913                                                                                              & 0.611                                                                                           & 0.805                                                                                                              \\ 
                 & EPAAE ($\zeta=2.5$)          & \textbf{0.839}                            & 0.422                               & \textbf{0.367}                      & \textbf{0.222}                       & \textbf{0.471}                      & \textbf{2.022}                      & 0.930                                                                                              & \textbf{0.647}                                                                                           & \textbf{0.827}                                                                                                     \\ 
                 & EPAAE ($\zeta=2.5,p=0.3$)    & 0.813                                     & 0.495                               & 0.276                               & 0.171                                & 0.355                               & 1.241                               & 0.906                                                                                              & 0.560                                                                                           & 0.770                                                                                                              \\ 
                 & EPAAE ($\zeta=2.5,p=0.1$)    & 0.827                                     & 0.440                               & 0.352                               & 0.215                                & 0.455                               & 1.931                               & \textbf{0.932}                                                                                     & 0.635                                                                                  & 0.818                                                                                                              \\
\hline
\end{tabular}
}
\caption{Quantitative results of TST experiments on NLI datasets}
\label{tab:disc}
\vspace{-.2in}
\end{table*}
\begin{table*}
\resizebox{1.0\textwidth}{!}{
\centering
\begin{tabular}{ccccccccccc} 
\hline
\textbf{Dataset} & \textbf{Model} & \multicolumn{1}{c}{\textbf{Naturalness}} & \multicolumn{1}{c}{\textbf{TST Acc.}} & \multicolumn{1}{c}{\textbf{BLEU-2}} & \multicolumn{1}{c}{\textbf{METEOR}} & \multicolumn{1}{c}{\textbf{ROUGE-L}} & \multicolumn{1}{c}{\textbf{CIDER}} & \multicolumn{1}{c}{\begin{tabular}[c]{@{}l@{}}\textbf{Embedding }\\\textbf{Average}\end{tabular}} & \multicolumn{1}{c}{\begin{tabular}[c]{@{}l@{}}\textbf{Vector}\\\textbf{~Extrema}\end{tabular}} & \multicolumn{1}{c}{\begin{tabular}[c]{@{}l@{}}\textbf{Greedy}\\\textbf{~Matching }\\\textbf{Score}\end{tabular}}  \\ 
\hline
\textbf{Voices}  & $\beta$-VAE ($ \beta=0.15$)         & 0.779                                     & 0.985                               & 0.144                               & 0.108                                & 0.204                               & 0.728                               & 0.765                                                                                              & 0.426                                                                                           & 0.593                                                                                                              \\ 
                 & LAAE ($\lambda_1=0.05$)        & 0.789                                     & 0.988                               & 0.095                               & 0.080                                & 0.162                               & 0.426                               & 0.746                                                                                              & 0.390                                                                                           & 0.560                                                                                                              \\ 
                 & DAAE ($p=0.3$)          & 0.783                                     & 0.981                               & 0.243                               & 0.182                                & 0.300                               & 1.488                               & 0.811                                                                                              & 0.523                                                                                           & 0.670                                                                                                              \\ 
                 & EPAAE ($\zeta=2.5$)          & \textbf{0.806}                            & \textbf{0.993}                               & 0.176                               & 0.134                                & 0.260                               & 0.919                               & 0.795                                                                                              & 0.468                                                                                           & 0.634                                                                                                              \\ 
                 & EPAAE ($\zeta=2.5,p=0.3$)    & 0.785                                     & 0.975                               & 0.174                               & 0.133                                & 0.237                               & 0.978                               & 0.777                                                                                              & 0.457                                                                                           & 0.616                                                                                                              \\ 
                 & EPAAE $(\zeta=2.5,p=0.1)$    & 0.796                                     & 0.991                      & \textbf{0.253}                      & \textbf{0.187}                       & \textbf{0.318}                      & \textbf{1.491}                      & \textbf{0.819}                                                                                     & \textbf{0.532}                                                                                  & \textbf{0.681}                                                                                                     \\ 
\hline
\textbf{PPR}     & $\beta$-VAE ($ \beta=0.15$)         & 0.735                                     & 0.949                               & 0.197                               & 0.202                                & 0.391                               & 1.291                               & 0.808                                                                                              & 0.536                                                                                           & 0.706                                                                                                              \\ 
                 & LAAE ($\lambda_1=0.05$)        & 0.748                                     & 0.940                               & 0.149                               & 0.143                                & 0.302                               & 0.846                               & 0.772                                                                                              & 0.467                                                                                           & 0.643                                                                                                              \\ 
                 & EPAAE ($p=0.3$)          & 0.730                                     & 0.948                               & 0.261                               & 0.279                                & 0.486                               & 1.797                               & 0.842                                                                                              & 0.615                                                                                           & 0.774                                                                                                              \\ 
                 & EPAAE ($\zeta=2.5$)          & \textbf{0.757}                            & 0.932                               & 0.282                               & 0.283                                & 0.510                               & 1.982                               & 0.857                                                                                              & 0.631                                                                                           & 0.789                                                                                                              \\ 
                 & EPAAE ($\zeta=2.5,p=0.3$)    & 0.728                                     & \textbf{0.955}                      & 0.236                               & 0.257                                & 0.448                               & 1.586                               & 0.829                                                                                              & 0.585                                                                                           & 0.749                                                                                                              \\ 
                 & EPAAE ($\zeta=2.5,p=0.1$)    & 0.747                                     & 0.938                               & \textbf{0.293}                      & \textbf{0.301}                       & \textbf{0.525}                      & \textbf{2.105}                      & \textbf{0.861}                                                                                     & \textbf{0.645}                                                                                  & \textbf{0.799}                                                                                                     \\ 
\hline
\textbf{Tenses}  & $\beta$-VAE ($ \beta=0.15$)         & 0.803                                     & 0.999                               & 0.110                               & 0.089                                & 0.198                               & 0.513                               & 0.766                                                                                              & 0.403                                                                                           & 0.583                                                                                                              \\ 
                 & LAAE ($\lambda_1=0.05$)        & \textbf{0.807}                            & \textbf{1.000}                      & 0.086                               & 0.072                                & 0.169                               & 0.388                               & 0.749                                                                                              & 0.377                                                                                           & 0.563                                                                                                              \\ 
                 & DAAE ($p=0.3$)          & 0.774                                     & 0.999                               & 0.263                               & 0.200                                & 0.363                               & 1.684                               & 0.822                                                                                              & 0.540                                                                                           & 0.686                                                                                                              \\ 
                 & EPAAE ($\zeta=2.5$)          & 0.798                                     & \textbf{1.000}                      & \textbf{0.358}                      & \textbf{0.274}                       & \textbf{0.483}                      & \textbf{2.305}                      & \textbf{0.861}                                                                                     & \textbf{0.631}                                                                                  & \textbf{0.757}                                                                                                     \\ 
                 & EPAAE ($\zeta=2.5,p=0.3$)    & 0.776                                     & 0.999                               & 0.302                               & 0.234                                & 0.408                               & 2.023                               & 0.834                                                                                              & 0.574                                                                                           & 0.711                                                                                                              \\ 
                 & EPAAE ($\zeta=2.5,p=0.1$)    & 0.785                                     & \textbf{1.000}                      & 0.339                               & 0.254                                & 0.456                               & 2.279                               & 0.852                                                                                              & 0.604                                                                                           & 0.739                                                                                                              \\
\hline
\end{tabular}
}
\caption{Quantitative results of TST experiments on Style-PTB datasets}
\label{tab:finegrained}
\vspace{-.1in}
\end{table*}
In this section, we look at the quantitative and quantitative results for the text style transfer task for four autoencoder models in seven datasets. We use the vector arithmetic method on latent space representations, inspired by \cite{inspired} where it showed that word embeddings learnt can capture linguistic relationships using simple vector arithmetic. Analogous to the standard example where ``King'' - ``Man'' + ``Woman'' $\approx$ ``Queen'', we manipulate an arbitrary sentence encoding $z_{x}$ of Style X to Style Y \footnote{Simply put, the difference of the means of all vectors in Style Y and X is computed and scaled by a factor $k$, then added to any arbitrary latent vector with style $Y$ to convert it to style $X$.}:
\begin{align}
        z_{x} =~ &  z_{y} ~+ k( \frac{1}{N_{y}}\sum_{i=0}^{N_{y}} z^i_{y} -  \frac{1}{N_{x}}\sum_{i=0}^{N_{x}} z^i_{x})
\end{align}

where $z^i_{y}$, $z^i_{y}$ denotes the latent vector of the $i^{th}$ sentence in style $y$ and $x$ respectively and $N_{x}$ and $N_{y}$ represent the number of encoding present in the corpus for style $x$ and $y$ respectively. $k$ is a scaling parameter used to control the style transfer strength. The resultant latent vector is passed through the decoder to produce the output sentence.

\subsection{Quantitative Analysis}
TST accuracy, content preservation and naturalness were computed on the converted sentences (shown in Table \ref{tab:sentiment}, \ref{tab:disc}, \ref{tab:finegrained}). Content preservation metrics can be found from Column 4 onwards. We consider two versions of the EPAAE i.e. only continuous embedding noise, continuous embedding noise + token deletion, and find that in some cases a mixture of both is required for optimal performance. We find that a slightly lowered value $p$ for the EPAAE combined with its optimal $\zeta$ parameter outperforms other models as well. 
\subsubsection{Sentiment TST}
Table \ref{tab:sentiment} summarises the results of TST on the Yelp dataset. On visual inspection, there appears to be a general tradeoff between TST\% and content preservation metrics. For example, the $\beta$-VAE achieves the best TST\% but suffers from bad content preservation capability. In this case, we observe that EPAAE ($\zeta=2.0,p=0.1)$ has the best content preservation capabilities across all metrics and achieves a reasonable tradeoff of TST\%=77.1 as well.
\subsubsection{Discourse TST}
Table \ref{tab:disc} summarises the results on four NLI datasets. Similar to the sentiment task, we see that EPAAEs have the best content preservation capabilities as well as the naturalness metric. It achieves this while achieving a comparable TST\% as well. $\beta$-VAE shows best TST\% but again suffers in content preservation. DAAE also display reasonable TST\% vs Content preservation tradeoffs but overall cannot match the tradeoffs achieved by EPAAEs. Human evaluations on the SNLI dataset (Table \ref{tab:humanevals}) confirm this as well. The TST\% achieved by any model in any task peaks at only 60.4\% compared to 81.9\% in sentiment task, a significant difference that hints at the fact that the Discourse TST task might be intrinsically more complex than sentiment style. Future work that aims to increase performance on the NLI task will be beneficial.

\subsubsection{Style-PTB TST}
Table \ref{tab:finegrained} summarises the results on three datasets from the Style-PTB benchmark. \cite{styleptb} produces a hierarchy of styles based on transfer difficulty measured by the average token-level hamming distance between the base and converted sentence. According to this hierarchy Tense inversion, PP removal/addition and Voice change are labelled in ascending order as easy, medium and hard respectively. Our results seem to partially validate this observation, in that the max TST\% is obtained on the Tenses dataset (100\%). Generally speaking, we also observe that TST has much better performance on fine-grained Style-PTB datasets than sentiment and discourse styles. Similar to before, the EPAAE shows best content preservation at competitive values of TST\% as well.

It is also noteworthy to consider the direction of style transfer, particularly in the case of complex styles such as Discourse styles present in NLI datasets. Results and analysis on direction-specific metrics for Discourse TST are presented in Appendix \ref{directionspecific}. 

\subsection{Qualitative Analysis}

\textbf{Samples of Output.}  For qualitative analysis, sample outputs by DAAE and EPAAE for the Yelp, SNLI and Tenses dataset are given in Table \ref{tab:examples1}. For a full list of qualitative examples on all datasets along with setup details, please refer to Appendix \ref{qualitativeexamples}. 



\textbf{Varying $k$ for Fine-Grained TST.} By varying the value of $k$ in Equation 5, it is possible to finely control the strength of Text Style transfer. We show examples of this for the Yelp (Table \ref{yelpfinegrained}) and the SNLI (Table \ref{snlifinegrained}) dataset, for the baseline DAAE ($p=0.3$) and the proposed model EPAAE. For the proposed EPAAE model, the best performing models (specifically in content preservation metrics) were chosen i.e. $zeta=2.0$, $p=0.1$ for the Yelp dataset and $zeta=2.5$ for the SNLI dataset. The chosen qualitative examples highlight the EPAAEs slight superiority in performing fine-grained TST compared to the baseline.
\begin{table*}
\resizebox{1.0\textwidth}{!}{
\centering
\begin{tabular}{ccc} 
\hline
               & \multicolumn{1}{c}{\textbf{DAAE ($p=0.3$)}}                           & \multicolumn{1}{c}{\textbf{EPAAE ($\zeta=2.0,p=0.1$)}}                            \\ 
\hline
\textbf{Input} & \textbf{the dog sits by a snowdrift . a dog out in the snow}         & \textbf{the dog sits by a snowdrift . a dog out in the snow}                 \\ 
\hline
$k=1$            & the dog stands in the snow . a dog stands in the snow                  & the dog sits under a fallen tree dog . a dog resting in the snow               \\ 
$k=1.5$          & the dog stands in the snow . a dog swims in the snow                   & the dog sits under a fallen tree dog . a dog resting in the snow               \\ 
$k=2$            & the dog stands in the snow . a dog swims in the snow                   & the dog sits under a fallen tree dog . a dog resting in the snow               \\ 
$k=2.5$          & the dog stands in the snow . a dog swims in the snow                   & the dog sits under a fallen tree dog . a dog resting in the snow at the porch  \\ 
$k=3$            & the dog leaps across the snow . a dog swims in the snow in the snow    & the dog sits under a fallen tree dog . a dog resting in the snow at the porch  \\ 
\hline
\textbf{Input} & \textbf{a man enjoys some extravagant artwork . a man is making art} & \textbf{a man enjoys some extravagant artwork . a man is making art}         \\ 
\hline
$k=-1$           & a man makes a strange art that says unk . a man is making clothing     & a man examines an art unk stall . a man is making art                          \\ 
$k=-1.5$         & a man makes a strange art that says unk . a man is making clothing     & a man examines an art unk by . a man is making art                             \\ 
$k=-2$           & a man makes a strange art that says unk . a man is making clothing     & a man examines an art unk by . a man is making art                             \\ 
$k=-2.5$         & a man makes a strange art that says art . a man is making clothing     & a man examines an art gallery unk . a man is making art                        \\ 
$k=-3$           & a man makes a strange art that says art . a man is making clothing     & a man examines an art gallery unk . a man is making art                        \\
\hline
\end{tabular}
}
\caption{Qualitative examples of Fine-grained TST on the SNLI Dataset (Entailment to Contradiction)}
\label{snlifinegrained}
\end{table*}

\textbf{Smooth Interpolation. }Sentence interpolation experiments are reported in Appendix \ref{interpolation}, in which latent space points in an interpolation along a specific direction are decoded to gauge the smoothness of the space and its ability to generate coherent sentences

\subsection{Human Evaluations}
Each human annotator was given a set of base sentences and asked to vote for which model produced the most appropriate corresponding style inverted sentences. Please refer to Appendix \ref{humandetails} for full details on the setup. Results are shown in Table \ref{tab:humanevals}. We observe that the proposed EPAAE model was overall more preferred across all three chosen datasets. This margin was most significant in the case of the SNLI dataset.
\begin{table}[ht]
\centering
\resizebox{0.5\textwidth}{!}{
\begin{tabular}{cccccc} 
\hline
\textbf{Dataset} & \multicolumn{1}{c}{\begin{tabular}[c]{@{}l@{}}\textbf{DAAE }\\\textbf{better}\end{tabular}} & \multicolumn{1}{c}{\begin{tabular}[c]{@{}l@{}}\textbf{EPAAE}\\\textbf{~better}\end{tabular}} & \multicolumn{1}{c}{\begin{tabular}[c]{@{}l@{}}\textbf{Both }\\\textbf{Good}\end{tabular}} & \multicolumn{1}{c}{\begin{tabular}[c]{@{}l@{}}\textbf{Both }\\\textbf{Bad}\end{tabular}} & \multicolumn{1}{c}{\begin{tabular}[c]{@{}l@{}}\textbf{No }\\\textbf{Agreement}\end{tabular}}  \\ 
\hline
\textbf{Yelp}    & 34                                                                                           & \textbf{47}                                                                                   & 16                                                                                         & 75                                                                                        & 28                                                                                             \\ 
\textbf{SNLI}    & 52                                                                                           & \textbf{90}                                                                                   & 4                                                                                          & 35                                                                                        & 19                                                                                             \\ 
\textbf{Tenses}  & 45                                                                                           & \textbf{59}                                                                                   & 17                                                                                         & 63                                                                                        & 16                                                                                             \\
\hline
\end{tabular}

}
\caption{Human Evaluations for Text Style Transfer on Yelp, SNLI and Tenses datasets. Bold indicates model with highest votes}
\label{tab:humanevals}
\end{table}

\section{Conclusion}
We introduce the ``Embedding Perturbed AAE'' or EPAAE and show that it best captures underlying style-based semantic features in the latent space in an unsupervised manner compared to its baselines. By inducing robust latent space organization through embedding perturbation in an unsupervised manner, we also demonstrate the possibility of fine-grained TST, where we can control the strength of the target style. Using a diverse set of datasets with varying formulations and complexities of style, we empirically that EPAAE performs overall best in the text style transfer task, particularly in its ability to preserve style-independent content across all datasets. 
\section{Future Work and Limitations}

\textbf{Regarding work in TST.} We wish to augment existing state of the art methods with embedding perturbation to check if doing so aids performance. We see degrading TST performance in the Entailment to Contradiction Task across all models. Future work will focus on methods to improve this task. \\
\textbf{Generally.} It is also interesting to further analyse the effects of embedding perturbation to latent representations and resultant properties. A theoretical analysis would be beneficial to cement the use of embedding perturbation in a more general setting. There also remain more important questions that need answering for, e.g. "What if you apply continuous perturbation to hidden states instead of embeddings?", "What is the relation between this type of perturbation and techniques like dropout?". We wish to explore these important questions in the future.
\section{Ethics Statement}
Any TST model can be used for nefarious purposes, e.g. performing a "Non-toxic to toxic" modification of text in a real-world setting and causing social harm. Therefore, it is important we keep in mind a code of ethics (e.g. \url{https://www.acm.org/code-of-ethics}) for usage, research and development in this type of research. We have made all our code open-source and provided all details of experimentation and implementation to the best of our knowledge. 

\section{Acknowledgements}
We would like to thank the reviewers whose feedback we believe substantially increased the quality of this work. We would like to thank the human annotators for their participation.

\bibliography{acl2020}
\bibliographystyle{acl_natbib}
\appendix
\renewcommand{\theHsection}{A\arabic{section}}
\renewcommand{\theHtable}{A\arabic{table}}
\renewcommand{\theHfigure}{A\arabic{figure}}

\textbf{\large{Appendix}} 
\renewcommand\thefigure{\thesection.\arabic{figure}}    
\setcounter{figure}{0}
\renewcommand\thetable{\thesection.\arabic{table}}    
\setcounter{table}{0}


\section{Additional details on Text Style Transfer Experiments}
\subsection{Model Architecture}
\label{modelarchitecture}
All baseline models were trained with all underlying architectures apart from their individual objective losses. Bi-directional GRUs were used for the encoder and decoder with input embeddings of size 300, a hidden representation of size 256 and a latent space of size 128. For all models using AAEs as the underlying autoencoder, the discriminator was a single-layered perceptron with 512 units. The ADAM optimiser with $\beta_1$, $\beta_2$ as 0.5, 0.999 and a learning rate of 0.001. All models were trained for 30 epochs as any more training steps caused the reconstruction loss to dominate and decrease overall performance on the TST task. All input perturbations were disabled in inference time.
\subsection{Hyperparameter Selection}
\label{hyperparameterselection}
 For the hyperparameters $p$, $\lambda_l$, and $\beta$ for the DAAE, LAAE and $\beta$-VAE, we fixed the values as 0.3, 0.05, 0.15 respectively. This decision was aligned with the results in \citet{daae}, which showed that these values produced the best reconstruction vs BLEU trade-off. We found this to be the case during subjective manual testing as well. $\lambda_{adv}$ was set to 10 for all models having AAEs as the underlying architecture. For EPAAE, we found that $\zeta \in [2.0,3.0]$ overall showed the best results across all datasets. Therefore a manual search around this range was conducted to determine the optimal $\zeta$ for each dataset. 

\subsection{Human Evaluations}
\label{humandetails}
TST outputs of two models, the baseline DAAE and the proposed EPAAE model, on the Yelp, SNLI and Tenses dataset were considered. The best performing EPAAE was chosen according to the results in Table \ref{tab:sentiment}, \ref{tab:disc} and \ref{tab:finegrained}, particularly with respect to the content preservation metrics (Since TST\% were similar across all models).
Two hundred sentences (hundred from each base style) were randomly sampled from the test split of each dataset and style inverted (with scaling factor $k=2$). Six hundred instances (each instance being a base and converted sentence pair) were equally split between three human evaluators. Each evaluator was given the task of labelling all two hundred instances from each dataset. The models were anonymous to evaluators and randomly named as "Model 1" and "Model 2". Each instance was to be labelled by an evaluator with four possible decision outcomes, i.e. "1 is best", "2 is best", "All are bad" and "All are good". For each instance, the majority of three votes from each of the three annotators were taken as the final decision for that corresponding instance. Instances without a majority were marked as "NA". The evaluation guidelines was formulated to consider which model a) successfully transferred the target style b) preserved the style-independent content and c) was overall fluent and grammatically coherent.

\section{Additional Experiments:}



\subsection{Sentence Interpolation in Latent Space - Qualitative Examples}
We perform sentence interpolation using DAAE and EPAAE, starting from the same input sentence, incrementally moving along the same direction in the latent space in five fixed-size steps. We see that both the baseline and proposed model are able to produce fluent and coherent sentences, indicative of a smoothly populated latent space. 
\label{interpolation}
\begin{table*}
\centering

\label{tab:snlinn}
\small

\begin{tabular}{p{7cm}p{7cm}} 
\hline
\textbf{two blond women are hugging one another the women are sleeping }                              & \textbf{two blond women are hugging one another the women are sleeping }                               \\ 
\hline
two blond women are hugging one another the women are sleeping                                        & two blond women are hugging one another the women are sleeping                                         \\ 
\hline
three dogs affectionately playing the dogs are sleeping                                               & three dogs affectionately playing the dogs are sleeping                                                \\ 
\hline
two dogs are playing and wrestling with each other two cats are chasing each other through the house  & two dogs are playing and wrestling with each other two cats are chasing each other through the house   \\ 
\hline
two men in wheelchairs crash and they reach for the ball the men are sleeping                         & two men in wheelchairs crash and they reach for the ball the men are sleeping                          \\ 
\hline
three dogs in different shades of brown and white biting and licking each other the dogs are fighting & three dogs in different shades of brown and white biting and licking each other the dogs are fighting  \\
\hline
\end{tabular}
\label{snlinn}
\caption{Sentence Interpolation on sentence from \textbf{SNLI} dataset (\textbf{Left:} EPAAE, \textbf{Right:} DAAE)}
\end{table*}
\begin{table*}
\centering
\small
\begin{tabular}{p{7cm}p{7cm}} 
\hline
\textbf{the report will follow five consecutive declines}                & \textbf{the report will follow five consecutive declines}                 \\ 
\hline
the report will follow five consecutive declines                         & the report will follow five consecutive declines                          \\ 
\hline
the report will follow five consecutive declines in full monthly figures & the report follows five consecutive declines                              \\ 
\hline
the report followed five consecutive declines                            & the report followed five consecutive declines                             \\ 
\hline
the report follows five consecutive declines                             & the report will follow five consecutive declines in full monthly figures  \\ 
\hline
the index hits its low 20297 off 2042 points                             & the report follows five consecutive declines in full monthly figures      \\
\hline
\end{tabular}
\label{tab:voicesnn}
\caption{Sentence Interpolation performed on the \textbf{Voices} dataset 
(\textbf{Left:} EPAAE, \textbf{Right:} DAAE)}
\end{table*}
\begin{table*}
\centering
\small
\begin{tabular}{p{7cm}p{7cm}} 
\hline
\textbf{the soviets had a world leading space program the guests noted} & \textbf{the soviets had a world leading space program the guests noted}  \\ 
\hline
the soviets had a world leading space program the guests noted          & the soviets had a world leading space program the guests noted           \\ 
\hline
the challengers had a big price advantage                               & the plant had a hairy stem that produced flowers and diminutive seeds    \\ 
\hline
the expansion set off a marketing war                                   & the japanese used 00 of the world s ivory                                \\ 
\hline
the uaw was seeking a hearing by the full 14 judge panel                & the diplomat added that mr krenz had several things going for him        \\ 
\hline
total return was price changes plus interest income                     & the japanese used 40 of the world s ivory                                \\
\hline
\end{tabular}
\label{tab:tensesnn}
\caption{Sentence Interpolation performed on the \textbf{Tenses} dataset (\textbf{Left:} EPAAE, \textbf{Right:} DAAE)}
\end{table*}

\subsection{Text Style Transfer}
\subsubsection{Qualitative Examples}
\label{qualitativeexamples}

\begin{table*} 
\centering
\small
\begin{tabular}{p{2cm}p{3cm}p{3cm}p{3cm}p{3cm}} 
\hline
\textbf{Yelp}     & \multicolumn{2}{c}{\textbf{Negative to Positive}}                                                                                                                         & \multicolumn{2}{c}{\textbf{Postive to Negative}}                                                                                                                                                   \\ 
\hline
Base              & i had high hopes and they simply could n't have fallen farther                                & probably wo n't stay here again                                            & the food is always flavorful and filling                                                  & food was excellent and service was fast                                                                 \\ 
\hline
Converted (DAAE)  & the italian flavors and amazing and all really impressive !                                   & certainly here !                                                           & the food is not cooked and it was appealing .                                             & food was not the waitress and was slow .                                                                \\ 
\hline
Converted (EPAAE) & i had great prices and everything is exquisite !                                              & definitely stay here !                                                     & the food is n't just extremely bland .                                                    & food was ok. service was slow and disappointed                                                          \\ 
\hline
\textbf{SNLI}     & \textbf{Entailment to Contradiction}                                                          &                                                                            & \textbf{Contradiction to Entailment}                                                      &                                                                                                         \\ 
\hline
Base              & a man and lady standing on a seesaw at a park . a man and woman are on a seesaw outside       & a dog is jumping through the water . tha animal is in the water            & a young child in a green shirt is on a carousel . the young child is wearing a blue shirt & three people shopping in an isle in a foreign grocery store . three people look at clothes in the mall  \\ 
\hline
Converted (DAAE)  & a man and a woman are on a seesaw at a beach. a man and woman are on a seesaw                 & a dog is running through the water the animal is running through the water & a young child in a blue shirt is wearing a red shirt and a blue hat a child is on a swing & three people in shopping carts in a shopping mall. three people shopping in a store                     \\ 
\hline
Converted (EPAAE) & a man and a woman are sitting on a seesaw at a park. a man and a woman are sitting on a couch & a dog is jumping through the water the animal is flying through the air    & a young child in a red shirt is on a green slide. the child is wearing a shirt            & three people shopping in a shopping mall in a foreign city. three people shopping in the mall           \\ 
\hline
\textbf{Tenses}   & \multicolumn{2}{c}{\textbf{Future to Past}}                                                                                                                               & \multicolumn{2}{c}{\textbf{Past to Future}}                                                                                                                                                        \\ 
\hline
Base              & miller brewing co and general motors will be included by clients                              & i will see a possibility of going to 2200 this month                       & spiegel was 80 controlled                                                                 & mr deaver had reopened a public relations business                                                      \\ 
\hline
Converted (DAAE)  & miller brewing co and general motors were included                                            & i was a math major but i was going at this                                 & 42 will be advanced                                                                       & mr breeden will not be public relations business will say                                               \\ 
\hline
Converted (EPAAE) & miller brewing co and general motors was included by clients                                  & i saw a possibility of going to 2200 this year                             & spiegel will be 80 controlled                                                             & mr deaver will have reopened a public relations business                                                \\
\hline
\end{tabular}
\label{tab:examples1}
\caption{Qualitative examples of TST task by DAAE and EPAAE on Yelp, SNLI and Tenses dataset}
\label{tab:examples1}
\end{table*}

\begin{table*}
\centering
\small
\begin{tabular}{cp{2cm}p{3.5cm}p{3.5cm}p{3.5cm}} 
\hline
\multicolumn{5}{c}{\textbf{DNLI}}                                                                                                                                                                                                                                                     \\ 
\hline
\multirow{2}{*}{\textbf{Models}} & \multicolumn{2}{c}{\textbf{Entailment to Contradiction}}                                                          & \multicolumn{2}{c}{\textbf{Contradiction to Entailment}}                                                                       \\ 
\cline{2-5}
                                 & \multicolumn{1}{c}{Example 1}                        & \multicolumn{1}{c}{Example 2}                             & \multicolumn{1}{c}{Example 1}                                 & \multicolumn{1}{c}{Example 2}                                 \\ 
\hline
\textbf{Base}                    & i work in a cubicle . i have a hectic job             & i have 3 kids . i am the proud parent of 2 boys and 1 girl & i work as a contractor for a cab company . i am a travel agent & i am a ballet dancer . i work as a trauma surgeon              \\ 
\hline
\textbf{DAAE}                    & i work as a cashier in a supermarket . i am a cashier & i have a wife and two kids . i am a proud us sailor        & i work for a online company . i do not have a job              & i am a ballet dancer . i work at home editing                  \\ 
\hline
\textbf{EPAAE}                   & i work in a cubicle . i am a skilled craftsman        & i have 3 kids . i am a proud mother of two                 & i work for a cab company . i have been working for many years  & i am a ballet dancer . i took ballet lessons when i was a kid  \\
\hline
\end{tabular}
\caption{TST task performed on DNLI dataset}
\label{tab:dnli}
\end{table*}

\begin{table*}
\centering
\small
\begin{tabular}{cp{3.1cm}p{3.1cm}p{3.1cm}p{3.1cm}} 
\hline
\multicolumn{5}{c}{\textbf{Scitail}}                                                                                                                                                                                                                                                                                                                                                                                                                                                                                                                                                                                                                                                                                                                                                                                           \\ 
\hline
\multirow{2}{*}{\textbf{Models}} & \multicolumn{2}{c}{\textbf{Entailment to Neutrality}}                                                                                                                                                                                                                                                                                                                                                                & \multicolumn{2}{c}{\textbf{Neutrality to Entailment}}                                                                                                                                                                                                                                                                                                                \\ 
\cline{2-5}
                                 & \multicolumn{1}{c}{Example 1}                                                                                                                                                        & \multicolumn{1}{c}{Example 2}                                                                                                                                                                     & \multicolumn{1}{c}{Example 1}                                                                                                                           & \multicolumn{1}{c}{Example 2}                                                                                                                                                                    \\ 
\hline
\textbf{Base}                    & most amphibians such as frogs live part of their lives on land and return to water to breed . frogs are amphibians that live part of the time in fresh water and live rest of the time on land & reptiles are found on every continent worldwide with the exception of the polar antarctica . modern members of the reptiles group live in many different habitats and are found on every continent except antarctica & respiratory rate is 40 150 breaths per minute . the normal respiratory rate per minute in adult humans is 12 18 breaths                                           & and was the goddess of earth and sky moon and sun . the sun and the moon appear to be about the same size in the sky because the moon is smaller in diameter and is closer to earth than the sun  \\ 
\hline
\textbf{DAAE}                    & most amphibians have dual habitats and the remainder on land and live in the first decade of fish and live in the first living things have the time of fresh water and live rest               & many species are the most common ancestor in the unk of the same species of birds and mammals have the same species of birds                                                                                         & cardiac output is the amount of breaths per minute the normal respiratory rate per minute the normal respiratory rate per minute in adult humans is 12 18 breaths & the sun and the moon is the closest planet to the sun and the moon is the closest planet to the sun the sun is the closest star                                                                   \\ 
\hline
\textbf{EPAAE}                   & most species live on land and breed on land frogs are amphibians that live part of the time to live in the ocean                                                                               & reptiles are found on every continent except antarctica the majority of species live in the pacific continent except antarctica and birds are found on every continent except antarctica                             & respiratory rate is a series of breaths per minute in the normal respiratory rate per minute in adult humans is 12 18 breaths                                     & earth and moon is about 400 times of the earth and that it is about 400 times of the sun and the moon appear to be about the same size in the sky because                                         \\
\hline
\end{tabular}
\caption{TST task performed on the SciTail dataset}
\label{tab:scitail}
\end{table*}

\begin{table*}
\centering
\small
\begin{tabular}{cp{3.1cm}p{3.1cm}p{3.1cm}p{3.1cm}} 
\hline
\multicolumn{5}{c}{\textbf{Voices}}                                                                                                                                                                                                                                                                                                   \\ 
\hline
\multirow{2}{*}{\textbf{Models}} & \multicolumn{2}{c}{\textbf{Active to Passive}}                                                                                   & \multicolumn{2}{c}{\textbf{Passive to Active}}                                                                                                                           \\ 
\cline{2-5}
                                 & \multicolumn{1}{c}{Example 1}                                  & \multicolumn{1}{c}{Example 2}                                  & \multicolumn{1}{c}{Example 1}                                                       & \multicolumn{1}{c}{Example 2}                                            \\ 
\hline
Base                             & he will make his remarks to a plo gathering in baghdad          & accounting problems will raise more knotty issues               & one stuck to old line business traditions while the change was embraced by the other & the only time it is had by the violin is right at the end                 \\ 
\hline
DAAE                             & he will be told by him as a sewage treatment plant he will say  & transportation services will be provided by accounting problems & outside i spotted two young exchange the company s own microprocessor said           & the soviets eavesdropping pays off however because the contract they say  \\ 
\hline
EPAAE                            & his remarks will be made by him from a plo gathering in baghdad & more knotty issues will be raised by accounting problems        & one stuck to old line business traditions while the other embraced the change        & the only time the violin has it s right to the end                        \\
\hline
\end{tabular}
\caption{TST task performed on Voices dataset}
\label{tab:voices}
\end{table*}

\begin{table*}
\centering
\small

\begin{tabular}{cp{3.1cm}p{3.1cm}p{3.1cm}p{3.1cm}} 
\hline
\multicolumn{5}{c}{\textbf{PP}}                                                                                                                                                                                                                                                                \\ 
\hline
\multirow{2}{*}{\textbf{Models}} & \multicolumn{2}{c}{\textbf{PP Removal}}                                                             & \multicolumn{2}{c}{\textbf{Information Addition}}                                                                                                     \\ 
\cline{2-5}
                                 & \multicolumn{1}{c}{Example 1}                      & \multicolumn{1}{c}{Example 2}                 & \multicolumn{1}{c}{Example 1}                                        & \multicolumn{1}{c}{Example 2}                                                 \\ 
\hline
\textbf{Base}                    & the problems will be magnified by the june killings & the senate will not vote on six lesser charges & membership will have since swelled                                    & the dispute will pit two groups                                                \\ 
\hline
\textbf{DAAE}                    & the problems will be                                & the senate will come                           & membership will have since swelled to between 20 of smaller creditors & the dispute will pit two groups of claimants against each other of each other  \\ 
\hline
\textbf{EPAAE}                   & the problems will be magnified                      & the senate will not vote                       & membership will have since swelled to at least 21 since friday        & the dispute will pit two groups of claimants against the two of japan          \\
\hline
\end{tabular}
\caption{TST task performed on PP dataset}
\label{tab:ppr}
\end{table*}
TST with scaling factor $k=2$ was performed on DNLI, Scitail datasets as seen in \ref{tab:dnli} and \ref{tab:scitail} respectively. Similarly, it was also performed on Voices and PP Removal datasets from the Style-PTB benchmark as shown in \ref{tab:voices} and \ref{tab:ppr} respectively. The proposed model for each dataset, was chosen to be the EPAAE with best performance in the content preservation metrics (shown in Table \ref{tab:sentiment}, \ref{tab:disc} and \ref{tab:finegrained}). Samples were specifically selected in which at least one of the models was able to generate the ideal, style converted sentence with near-perfect content preservation and coherence. This was done by human evaluators, where both the models were anonymized.

\begin{table*}
\resizebox{1.0\textwidth}{!}{
\centering
\begin{tabular}{ccc} 
\hline
               & \multicolumn{1}{c}{\textbf{DAAE ($p=0.3$)}}                             & \multicolumn{1}{c}{\textbf{EPAAE ($\zeta=2.0,p=0.1$)}}                       \\ 
\hline
\textbf{Input} & \textbf{very disappointed they ran out of average things in the menu} & \textbf{very disappointed they ran out of average things in the menu}  \\ 
\hline
$k=1$            & very nice and ran out of the food in !                                & very disappointed and all in out of the menu .                         \\ 

$k=1$           & very nice and varied with the quality menu !                          & very disappointed and all great flavors in the menu .                  \\ 

$k=2$            & very nice and varied selection of food !                              & very prepared and great flavors in the menu .                          \\ 

$k=2.5$          & very nice locally and varied menu ! !                                 & very delicious flavors and great flavors in the                        \\ 

$k=3$            & very great selection and great seafood !                              & very delicious flavors and great fish selections .                     \\ 
\hline
\textbf{Input} & \textbf{food was excellent and service was fast}                      & \textbf{food was excellent and service was fast}                       \\ 
\hline
$k=-1$           & food was excellent and the service was fast                           & food was excellent and service was fast .                              \\ 

$k=-1.5$         & food was excellent and the service was fast .                         & food was excellent and service was fine but service                    \\ 

$k=-2$           & food was not the waitress and was slow .                              & food was undercooked and service was fine but delivery .               \\ 

$k=-2.5$         & it was not the waitress , but was slow .                              & food was undercooked and service was not fine at food .                \\ 

$k=-3$           & it was not ignored the service , but was slow .                       & !                                                                      \\
\hline

\end{tabular}
}
\caption{Qualitative examples of Fine-grained TST on the Yelp Dataset}
\label{yelpfinegrained}
\end{table*}
\subsubsection{Direction specific Discourse Transfer metrics}
\label{directionspecific}
Particularly in the case of Discourse Style, it is natural to speculate that the difficulty of style transfer might be sensitive to the direction i.e. is it Contradiction/Neutrality to Entailment or vice versa. 
Intuitively, this makes sense as the Entailment to Contradiction/Neutrality tasks can be achieved simply by randomly editing either the subject or predicate or both, in any one sentence, to trigger a contradiction/neutrality between the two. However, in the reverse task, the edited part must be carefully chosen to precisely match the context of the other sentence to trigger entailment.
\\\\
To analyze this, direction-specific quantitative metrics for Discourse TST are conducted for the SNLI (\ref{tab:snlidir}), DNLI (\ref{tab:dnlidir}) and SciTail (\ref{tab:scitaildir}) datasets. We notice a disparity in performances in fact, does exist, mainly highlighted by the differences in the TST\% metric. This sensitivity to direction is present in all models across all datasets but is most significant in the SNLI dataset in which TST\% goes as low as 20.6\% for the Contradiction to Entailment task and as high as 83.7\% for the opposite task. Future work can focus on trying to specifically improve the Contradiction to Entailment task, as doing so will be a measure of a model's ability to detect and carefully align the content of one sentence to match another.
\begin{figure*}
    \centering
    \includegraphics[scale=0.4, trim=25 25 25 25, clip]{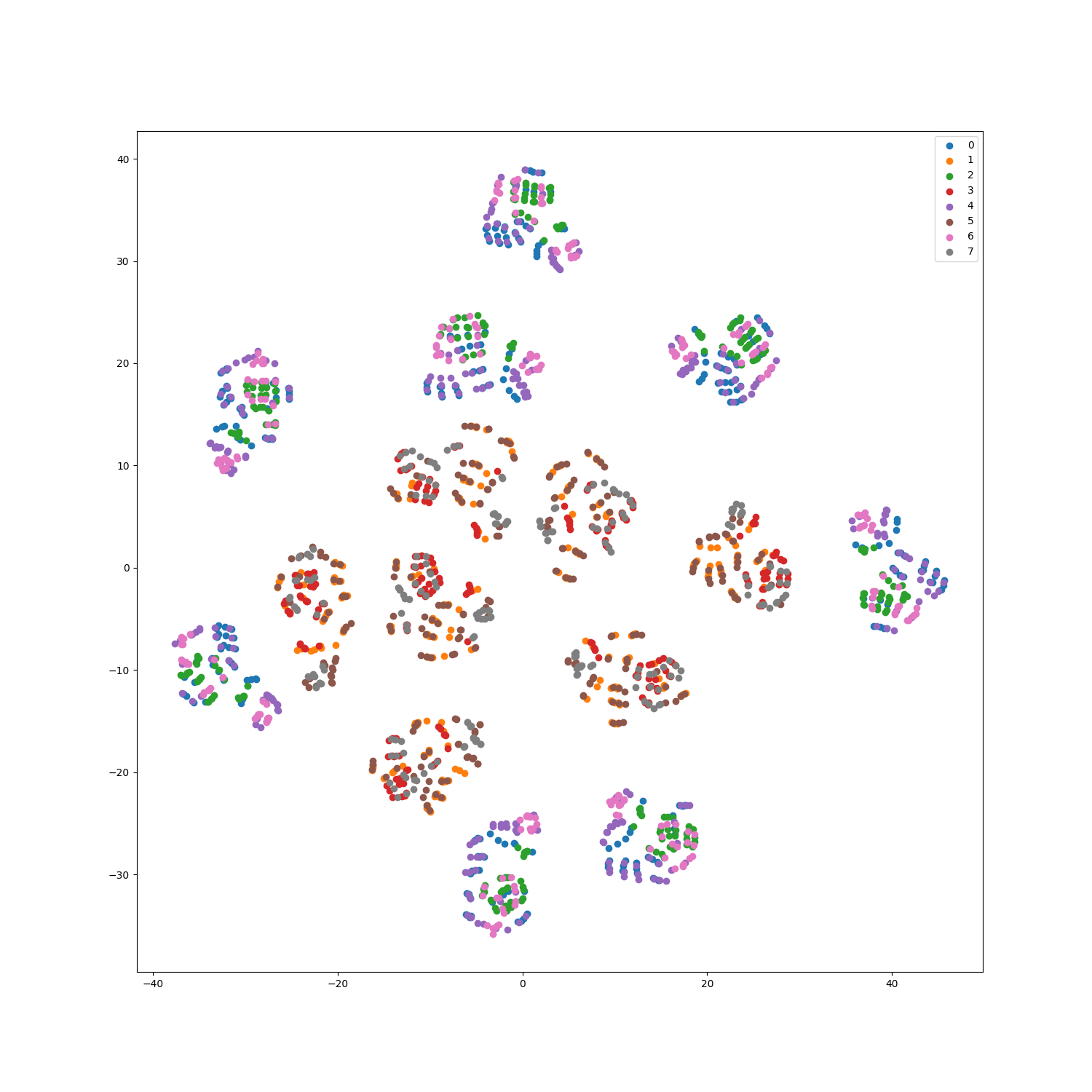}
    \caption{A TSNE plot of encoded latent vectors for all 1950 sentences in the Toy Dataset for the DAAE model.}
    \label{fig:fig1big}
    \vspace{-0.2in}
\end{figure*}

\begin{figure*}
    \centering
    \includegraphics[scale=0.4, trim=25 25 25 25, clip]{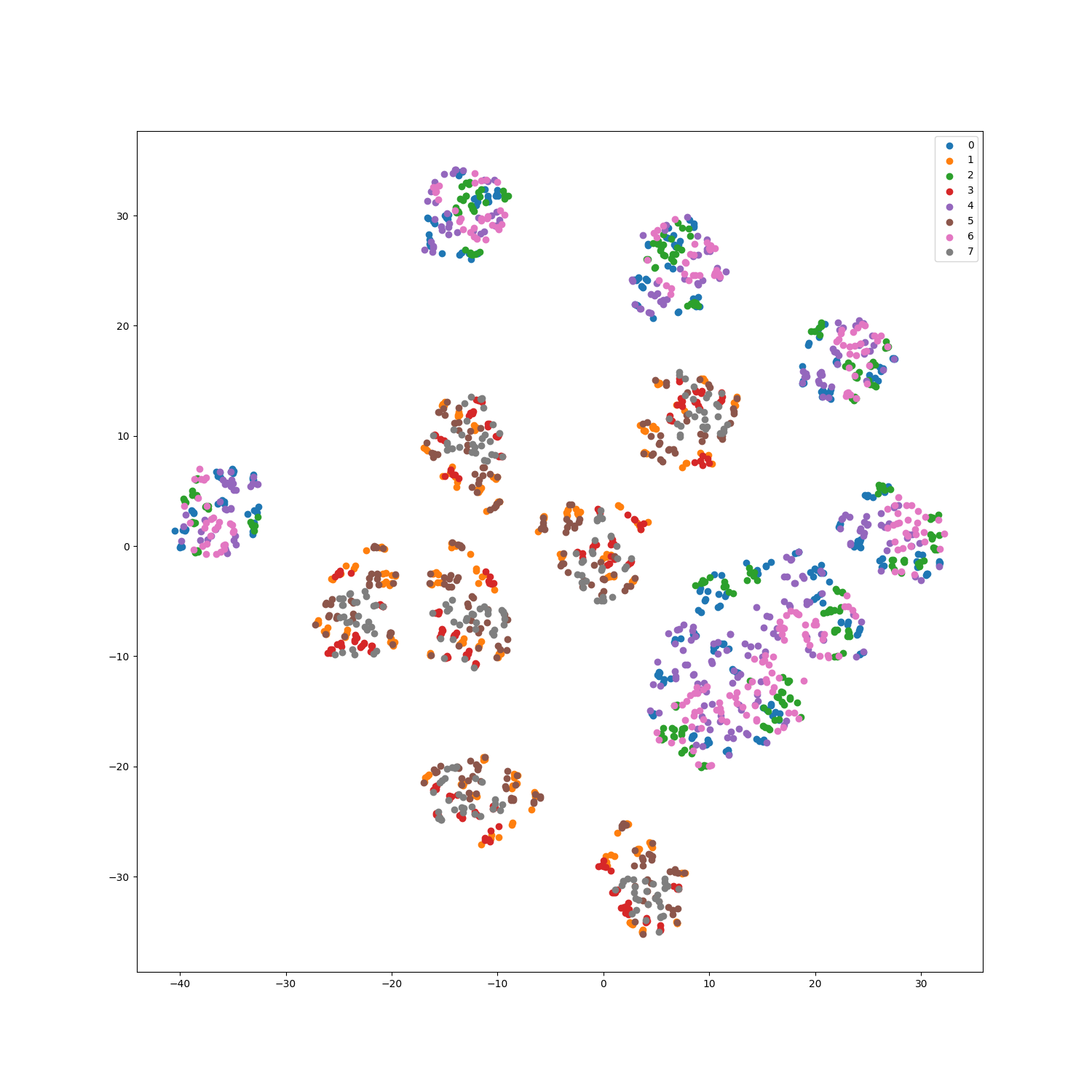}
    \caption{A TSNE plot of encoded latent vectors for all 1950 sentences in the Toy Dataset for the EPAAE model.}
    \label{fig:fig2big}
    \vspace{-0.2in}
\end{figure*}

\begin{table*}
\resizebox{1.0\textwidth}{!}{
\centering
\begin{tabular}{ccccccccccc} 
\hline
\textbf{Dataset}       & \textbf{Model} & \multicolumn{1}{c}{\textbf{Naturalness}} & \multicolumn{1}{c}{\textbf{TST Acc.}} & \multicolumn{1}{c}{\textbf{BLEU-2}} & \multicolumn{1}{c}{\textbf{METEOR}} & \multicolumn{1}{c}{\textbf{ROUGE}-\textbf{L}} & \multicolumn{1}{c}{\textbf{CIDEr}} & \multicolumn{1}{c}{\begin{tabular}[c]{@{}l@{}}\textbf{Embedding }\\\textbf{Average}\end{tabular}} & \multicolumn{1}{c}{\begin{tabular}[c]{@{}l@{}}\textbf{Vector }\\\textbf{Extrema}\end{tabular}} & \multicolumn{1}{c}{\begin{tabular}[c]{@{}l@{}}\textbf{Greedy }\\\textbf{Matching }\\\textbf{Score}\end{tabular}}  \\ 
\hline
\textbf{SNLI (E to C)} & $\beta$-VAE ($\beta=0.15$)         & \textbf{0.982}                            & 0.806                               & 0.438                                & 0.270                                & 0.555                                          & 2.455                               & 0.950                                                                                              & 0.630                                                                                           & 0.823                                                                                                              \\ 

                       & LAAE ($\lambda_{l}=0.05$)        & 0.947                                     & \textbf{0.837}                      & 0.036                                & 0.050                                & 0.160                                          & 0.027                               & 0.818                                                                                              & 0.320                                                                                           & 0.591                                                                                                              \\ 

                       & DAAE ($p=0.3$)          & 0.979                                     & 0.745                               & 0.453                                & 0.275                                & 0.562                                          & 2.410                               & 0.954                                                                                              & 0.679                                                                                           & 0.854                                                                                                              \\ 

                       & EPAAE ($\zeta=2.5$)          & 0.975                                     & 0.743                               & \textbf{0.524}                       & \textbf{0.312}                       & \textbf{0.641}                                 & \textbf{3.266}                      & \textbf{0.956}                                                                                     & 0.682                                                                                           & 0.856                                                                                                              \\ 

                       & EPAAE ($\zeta=2.5$, $p=0.3$)    & 0.975                                     & 0.749                               & 0.497                                & 0.300                                & 0.606                                          & 2.845                               & \textbf{0.956}                                                                                     & \textbf{0.698}                                                                                  & \textbf{0.866}                                                                                                     \\ 

                       & EPAAE ($\zeta=2.5$,$p=0.1$)    & \textbf{0.982}                            & 0.749                               & 0.405                                & 0.256                                & 0.522                                          & 1.983                               & 0.951                                                                                              & 0.648                                                                                           & 0.838                                                                                                              \\ 
\hline
\textbf{SNLI (C to E)} & $\beta$-VAE ($\beta=0.15$)         & \textbf{0.985}                            & 0.239                               & 0.411                                & 0.244                                & 0.527                                          & 2.121                               & 0.947                                                                                              & 0.599                                                                                           & 0.805                                                                                                              \\ 

                       & LAAE ($\lambda_l=0.05$)        & 0.945                                     & 0.206                               & 0.035                                & 0.052                                & 0.157                                          & 0.025                               & 0.829                                                                                              & 0.316                                                                                           & 0.594                                                                                                              \\ 

                       & DAAE ($p=0.3$)          & 0.982                                     & 0.292                               & 0.378                                & 0.233                                & 0.506                                          & 1.714                               & 0.946                                                                                              & 0.622                                                                                           & 0.823                                                                                                              \\ 

                       & EPAAE ($\zeta=2.5$),         & 0.979                                     & 0.227                               & \textbf{0.488}                       & \textbf{0.289}                       & \textbf{0.615}                                 & \textbf{2.860}                      & \textbf{0.956}                                                                                     & \textbf{0.659}                                                                                  & \textbf{0.843}                                                                                                     \\ 

                       & EPAAE ($\zeta=2.5$,$p=0.1$)    & 0.982                                     & 0.271                               & 0.425                                & 0.262                                & 0.557                                          & 2.140                               & 0.951                                                                                              & 0.650                                                                                           & 0.842                                                                                                              \\ 

                       & EPAAE ($\zeta=2.5$,$p=0.3$)    & 0.984                                     & \textbf{0.277}                      & 0.370                                & 0.225                                & 0.486                                          & 1.582                               & 0.945                                                                                              & 0.603                                                                                           & 0.815                                                                                                              \\
\hline

\end{tabular}
}
\caption{Direction-wise TST metrics for the SNLI dataset. "Entailment" and "Contradiction" are denoted by "E" and "C" respectively.}
\label{tab:snlidir}
\end{table*}
\begin{table*}
\resizebox{1.0\textwidth}{!}{
\centering

\begin{tabular}{ccccccccccc} 
\hline
\textbf{Dataset}       & \textbf{Model} & \multicolumn{1}{c}{\textbf{Naturalness}} & \multicolumn{1}{c}{\textbf{TST Acc.}} & \multicolumn{1}{c}{\textbf{BLEU-2}} & \multicolumn{1}{c}{\textbf{METEOR}} & \multicolumn{1}{c}{\textbf{ROUGE-L}} & \multicolumn{1}{c}{\textbf{CIDEr}} & \multicolumn{1}{c}{\begin{tabular}[c]{@{}l@{}}\textbf{Embedding }\\\textbf{Average}\end{tabular}} & \multicolumn{1}{c}{\begin{tabular}[c]{@{}l@{}}\textbf{Vector }\\\textbf{Extrema}\end{tabular}} & \multicolumn{1}{c}{\begin{tabular}[c]{@{}l@{}}\textbf{Greedy }\\\textbf{Matching }\\\textbf{Score}\end{tabular}}  \\ 
\hline
\textbf{DNLI (E to C)} & $\beta$-VAE ($\beta=0.15$)         & 0.922                                     & 0.645                               & 0.241                                & 0.142                                & 0.350                                 & 0.619                               & 0.902                                                                                              & 0.500                                                                                           & 0.720                                                                                                              \\ 

                       & LAAE ($\lambda_l=0.05$)        & 0.923                                     & 0.667                               & 0.203                                & 0.127                                & 0.314                                 & 0.376                               & 0.893                                                                                              & 0.473                                                                                           & 0.701                                                                                                              \\ 

                       & DAAE ($p=0.3$)          & \textbf{0.937}                            & \textbf{0.687}                      & 0.399                                & 0.223                                & 0.496                                 & 2.128                               & 0.925                                                                                              & 0.593                                                                                           & 0.790                                                                                                              \\ 

                       & EPAAE ($\zeta=2.5$)          & 0.934                                     & 0.536                               & \textbf{0.441}                       & \textbf{0.252}                       & \textbf{0.554}                        & 2.694                               & \textbf{0.941}                                                                                     & \textbf{0.636}                                                                                  & \textbf{0.818}                                                                                                     \\ 

                       & EPAAE ($\zeta=2.5$,$p=0.1$)    & 0.927                                     & 0.638                               & 0.433                                & 0.243                                & 0.533                                 & \textbf{2.732}                      & 0.934                                                                                              & 0.617                                                                                           & 0.809                                                                                                              \\ 

                       & EPAAE ($\zeta=2.5$,$p=0.3$)    & 0.923                                     & 0.665                               & 0.390                                & 0.218                                & 0.492                                 & 2.048                               & 0.925                                                                                              & 0.590                                                                                           & 0.790                                                                                                              \\ 
\hline
\textbf{DNLI (C to E)} & $\beta$-VAE ($\beta=0.15$)         & \textbf{0.948}                            & 0.516                               & 0.245                                & 0.146                                & 0.363                                 & 0.689                               & 0.908                                                                                              & 0.507                                                                                           & 0.724                                                                                                              \\ 

                       & LAAE ($\lambda_l=0.05$)        & 0.932                                     & \textbf{0.546}                      & 0.208                                & 0.129                                & 0.323                                 & 0.391                               & 0.900                                                                                              & 0.477                                                                                           & 0.707                                                                                                              \\ 

                       & DAAE ($p=0.3$)          & 0.930                                     & 0.514                               & 0.433                                & 0.251                                & 0.547                                 & 2.411                               & 0.943                                                                                              & 0.639                                                                                           & 0.817                                                                                                              \\ 

                       & EPAAE ($\zeta=2.5$)          & 0.934                                     & 0.492                               & 0.438                                & 0.253                                & 0.560                                 & 2.451                               & 0.945                                                                                              & 0.635                                                                                           & 0.818                                                                                                              \\ 

                       & EPAAE ($\zeta=2.5$,$p=0.1$)    & 0.928                                     & 0.519                               & \textbf{0.482}                       & \textbf{0.277}                       & \textbf{0.585}                        & \textbf{3.092}                      & \textbf{0.947}                                                                                     & \textbf{0.654}                                                                                  & \textbf{0.830}                                                                                                     \\ 

                       & EPAAE ($\zeta=2.5$,$p=0.3$)    & 0.945                                     & 0.490                               & 0.403                                & 0.233                                & 0.520                                 & 2.081                               & 0.940                                                                                              & 0.621                                                                                           & 0.806                                                                                                              \\
\hline
\end{tabular}
}

\caption{Direction-wise TST metrics for the DNLI dataset. "Entailment" and "Contradiction" are denoted by "E" and "C" respectively.}
\label{tab:dnlidir}
\end{table*}
\begin{table*}
\resizebox{1.0\textwidth}{!}{
\centering
\begin{tabular}{ccccccccccc} 
\hline
\textbf{Dataset}          & \textbf{Model} & \multicolumn{1}{c}{\textbf{Naturalness}} & \multicolumn{1}{c}{\textbf{TST Acc.}} & \multicolumn{1}{c}{\textbf{BLEU}-\textbf{2}} & \multicolumn{1}{c}{\textbf{METEOR}} & \multicolumn{1}{c}{\textbf{ROUGE-L}} & \multicolumn{1}{c}{\textbf{CIDEr}} & \multicolumn{1}{c}{\begin{tabular}[c]{@{}l@{}}\textbf{Embedding }\\\textbf{Average}\end{tabular}} & \multicolumn{1}{c}{\begin{tabular}[c]{@{}l@{}}\textbf{Vector }\\\textbf{Extrema}\end{tabular}} & \multicolumn{1}{c}{\begin{tabular}[c]{@{}l@{}}\textbf{Greedy }\\\textbf{Matching }\\\textbf{Score}\end{tabular}}  \\ 
\hline
\textbf{SciTail (E to N)} & $\beta$-VAE ($\beta=0.15$)         & 0.783                                     & 0.672                               & 0.086                                         & 0.060                                & 0.163                                 & 0.106                               & 0.799                                                                                              & 0.376                                                                                           & 0.640                                                                                                              \\ 

                          & LAAE ($\lambda_1=0.05$)        & 0.838                                     & 0.550                               & 0.173                                         & 0.111                                & 0.261                                 & 0.624                               & 0.870                                                                                              & 0.459                                                                                           & 0.695                                                                                                              \\ 

                          & DAAE ($p=0.3$)          & 0.804                                     & 0.514                               & 0.300                                         & 0.185                                & 0.383                                 & 1.375                               & 0.918                                                                                              & 0.579                                                                                           & 0.792                                                                                                              \\ 

                          & EPAAE ($\zeta=2.5$)          & \textbf{0.842}                            & 0.479                               & \textbf{0.351}                                & \textbf{0.210}                       & \textbf{0.445}                        & \textbf{1.941}                      & \textbf{0.931}                                                                                     & \textbf{0.618}                                                                                  & \textbf{0.816}                                                                                                     \\ 

                          & EPAAE ($\zeta=2.5,p=0.1$)    & 0.834                                     & 0.493                               & 0.328                                         & 0.198                                & 0.421                                 & 1.775                               & 0.929                                                                                              & 0.601                                                                                           & 0.803                                                                                                              \\ 

                          & EPAAE ($\zeta=2.5,p=0.3$)    & 0.829                                     & \textbf{0.571}                      & 0.234                                         & 0.149                                & 0.314                                 & 0.905                               & 0.899                                                                                              & 0.517                                                                                           & 0.747                                                                                                              \\ 
\hline
\textbf{SciTail (N to E)} & $\beta$-VAE ($\beta=0.15$)         & 0.748                                     & 0.397                               & 0.103                                         & 0.069                                & 0.183                                 & 0.178                               & 0.792                                                                                              & 0.399                                                                                           & 0.647                                                                                                              \\ 

                          & LAAE ($\lambda_1=0.05$)        & 0.823                                     & \textbf{0.408}                      & 0.219                                         & 0.138                                & 0.314                                 & 0.881                               & 0.878                                                                                              & 0.519                                                                                           & 0.729                                                                                                              \\ 

                          & DAAE ($p=0.3$)          & 0.772                                     & 0.352                               & 0.349                                         & 0.213                                & 0.444                                 & 1.745                               & 0.909                                                                                              & 0.644                                                                                           & 0.817                                                                                                              \\ 

                          & EPAAE ($\zeta=2.5$)          & \textbf{0.833}                            & 0.327                               & \textbf{0.384}                                & \textbf{0.233}                       & \textbf{0.497}                        & \textbf{2.104}                      & 0.929                                                                                              & \textbf{0.676}                                                                                  & \textbf{0.839}                                                                                                     \\ 

                          & EPAAE ($\zeta=2.5$,$p=0.1$)    & 0.816                                     & 0.351                               & 0.375                                         & 0.231                                & 0.488                                 & 2.087                               & \textbf{0.934}                                                                                     & 0.669                                                                                           & 0.834                                                                                                              \\ 

                          & EPAAE ($\zeta=2.5$,$p=0.3$)    & 0.786                                     & 0.368                               & 0.317                                         & 0.194                                & 0.397                                 & 1.577                               & 0.913                                                                                              & 0.603                                                                                           & 0.792                                                                                                              \\
\hline
\end{tabular}
}

\caption{Direction-wise TST metrics for the SciTail dataset. "Entailment" and "Neutrality" are denoted by "E" and "N" respectively }
\label{tab:scitaildir}
\end{table*}

\section{Details on Datasets}
\label{app:datasets}
Here we provide some additional details of all the datasets used in this work.

\textbf{Complexity of Styles in Datasets:} As discussed in Section \ref{sub:datasets}, we consider three tasks- sentiment, discourse and fine-grained text style transfer. As prepossessing, we remove non-essential special characters and lowercase all sentences. Except for the Yelp dataset, no pruning is done based on sentence length. The vocab size during training was limited at 25k  unless mentioned otherwise. 

\textbf{Sentiment Style Datasets: } We use the preprocessed version from \cite{shen} of the Yelp dataset. It contains 200k, 10k, 10k sentences in the train, dev and test split respectively. The sentiment labels (positive, negative) were considered as style.

\textbf{Discourse Style Datasets: }We used three NLI datasets - SNLI, DNLI, and Scitail. Each instance in the SNLI dataset \cite{snli} consists of two sentences. These sentences either contradict, entail (agree) or are neutral towards each other. The resultant dataset contained 341k, 18k, 18k in the train, dev, test splits respectively. The DNLI dataset \cite{dnli} consists of contradiction, entailment and neutrality labelled instances instead in the form of a first-person dialogue like representation. The dataset contains 208k, 11k, 11k sentences in train, dev and test respectively. Scitail \cite{scitail} is an entailment dataset created from multiple-choice science exams and the web, in a two-sentence format similar to SNLI and DNLI. The first sentence is formed from a question and answer pair from science exams and the second sentence is either a supporting (entailment) or non-supporting (neutrality) premise obtained from the internet. The dataset contained 24k, 1.3k, 1.3k sentences in the train, dev, test splits respectively.
For SNLI and DNLI,  all instances with the ``neutrality'' label were removed. Style transfer task performed from ``contradict'' to ``entail'' and vice versa. For SciTail \cite{scitail}, the transfer task was from ``neutral'' to ``entail'' and vice versa.

\textbf{Fine-grained Style Datasets: }We used Style-PTB dataset \cite{styleptb} for fine-grained style. It consists of 21 styles/labels with themes ranging from syntax, lexical, semantic and thematic transfers as well as compositional transfers which consist of transferring more than one of the aforementioned fine-grained styles. To check whether the EPAAE can capture fine-grained styles better by leveraging its better organized latent space, we make use of three styles i.e. Tenses, Voices (Active or Passive) and Syntactic PP tag removal (PPR). In the Tenses dataset, each sentence is labelled with "Present", "Past" and "Future". The Voices dataset contains "Active" and "Passive" voices labels and the PPR dataset contains "PP removed" and "PP not removed" labels. The resultant sizes of the test, dev, test splits were 71k,8.8k,8.8k (tenses), 90k,11k,11k (PPR) and 44k,5.5k,5.5k (voices).

\section{More details on related work}
\label{autoencoderwork}
\citet{bowman} extend Variational Auto-Encoders \cite{vae} for text generation and address the posterior collapse problem wherein the decoder completely ignores the latent channel leading to poor generation. Adversarial auto-encoders \cite{aae} substitute the KL loss with an adversarial approach to enforce the latent Gaussian prior. On the other hand, AAEs are shown to naturally avoid the posterior collapse problem and promote strong coupling between the encoder and decoder. Wasserstein autoencoders \cite{wasser} introduce a family of regularized autoencoders that learn a flexible prior by solving an optimal transport problem to match P(z) and Q(z) using an adversarial approach. Adversarially regularised autoencoders (ARAEs) follow the Wasserstein Autoencoder framework to learn a learnt prior unlike AAEs, which assume the prior to be a fixed standard Gaussian distribution.

\section{Computational Expense and Infrastructure used}
\label{infra}
The most parameter-heavy EPAAE model was from the SNLI dataset and we therefore report statistics for this model.
The model has 13 million parameters and each epoch took approximately 60 seconds to train on an Nvidia V100-SMX2 GPU and an Intel(R) Xeon(R) E5-2698 CPU. For complete details, please refer to the log.txt files present in each model's directory present in \url{https://github.com/sharan21/EPAAE}.

\end{document}